%% file: acl_latex.tex
\pdfoutput=1

\documentclass[11pt]{article}

\usepackage[final]{acl}

\usepackage{times}
\usepackage{latexsym}
\usepackage[T1]{fontenc}

\usepackage[utf8]{inputenc}

\usepackage{microtype}

\usepackage{inconsolata}

\usepackage{amsmath}
\usepackage{amssymb}  
\usepackage{graphicx}
\usepackage{xspace}
\usepackage{makecell}
\usepackage{adjustbox}
\usepackage{comment}
\usepackage{booktabs}
\usepackage{multirow}
\usepackage{caption}
\usepackage{subcaption,comment}
\newcommand\tf[1]{\textbf{#1}}

\newcommand{\model}{\textsc{ASRank}\xspace}
\title{ASRank: Zero-Shot Re-Ranking with Answer Scent for Document Retrieval}

\author{
    \textbf{Abdelrahman Abdallah, Jamshid Mozafari, Bhawna Piryani, Adam Jatowt} \\
    University of Innsbruck \\
    \texttt{\{abdelrahman.abdallah, jamshid.mozafari, bhawna.piryani,} \\
    \texttt{ adam.jatowt\}@uibk.ac.at}
}

\begin{document}
\maketitle
\begin{abstract}
Retrieval-Augmented Generation (RAG) models have drawn considerable attention in modern open-domain question answering. The effectiveness of RAG depends on the quality of the top retrieved documents. However, conventional retrieval methods sometimes fail to rank the most relevant documents at the top. In this paper, we introduce \model\footnote{\url{https://github.com/DataScienceUIBK/rankify}}, a new re-ranking method based on scoring retrieved documents using zero-shot answer scent which relies on a pre-trained large language model to compute the likelihood of the document-derived answers aligning with the answer scent.  Our approach demonstrates marked improvements across several datasets, including NQ, TriviaQA, WebQA, ArchivalQA, HotpotQA, and Entity Questions. Notably, \model increases Top-1 retrieval accuracy on NQ from $19.2\%$ to $46.5\%$ for MSS and $22.1\%$ to $47.3\%$ for BM25.  It also shows strong retrieval performance on several datasets compared to state-of-the-art methods (47.3 Top-1 by \model vs 35.4 by UPR by BM25).
\end{abstract}

\input{sections/introduction}

\input{sections/methods}

\input{sections/experimental_setup}
\input{sections/experimental_result}

\input{sections/related_work}

\vspace{-1mm}

\section{Conclusions}

In this paper, we introduced \model, a novel zero-shot re-ranking method that leverages the concept of answer scent to improve document retrieval for open-domain question answering and information retrieval tasks. Through comprehensive experiments on diverse datasets, including both open-domain and document-centric benchmarks like BEIR and TREC, we demonstrated that \model consistently outperforms unsupervised and supervised baselines. Our approach made significant improvements in Top-1, Top-5, and Top-10 retrieval accuracy, particularly when integrated with LLMs like GPT-3.5 and Llama-70B. Moreover, the cost-effectiveness and computational efficiency of \model, compared to other large-scale re-ranking methods, underscore its practical utility. 


\section*{Limitations}
While \model demonstrates significant improvements in document re-ranking with the incorporation of answer scent, there are several limitations that warrant discussion:
\begin{enumerate}
    \item The computational cost associated with \model increases with the number of documents due to the need to compute the score based on the answer scent with the answer generated from each document.
    \item The effectiveness and consistency of \model are contingent upon the specific pre-trained language models used for generating the answer scent. Variations in these models, due to different training data or updates, can introduce biases and affect the stability of the re-ranking outcomes.
    \item \model's performance heavily depends on the quality of the initial retrieval phase.
\end{enumerate}

\bibliography{custom}

\appendix

\input{sections/Appendix}

\end{document}

%% file: sections/introduction.tex
\section{Introduction}

\begin{figure}[t!]
\centering
\includegraphics[width=0.40\textwidth]{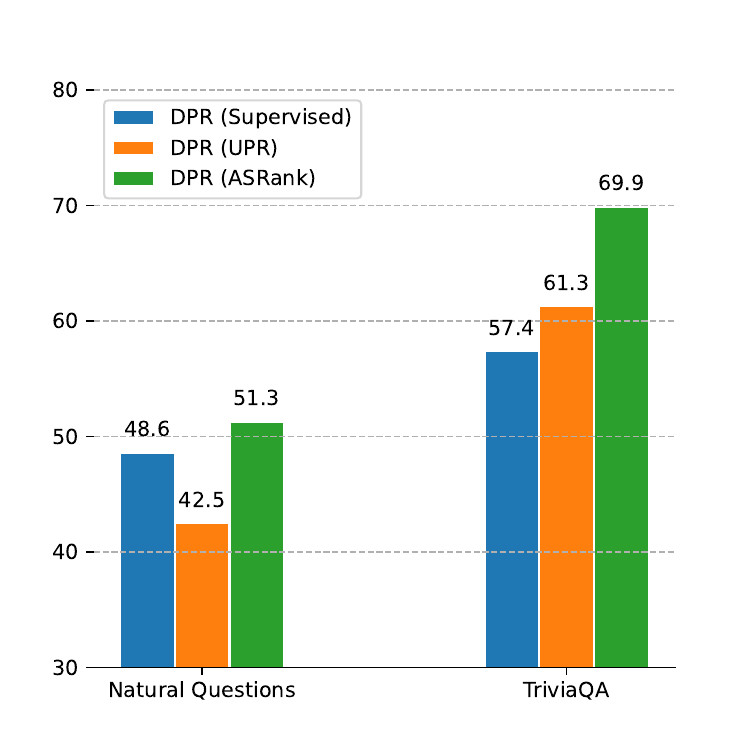}
\caption{
After re-ranking the top 1,000 passages retrieved by DPR~\cite{karpukhin-etal-2020-dense} with \model, our method surpasses the performance of strong unsupervised models like UPR \cite{sachan2022improving} on the Natural Questions and TriviaQA datasets.
}
\label{fig:intro}
\end{figure}

Document retrieval is a core sub-task in many NLP problems, including open-domain question answering (ODQA), where a document is retrieved and then read to answer an input query. This process tries to find the most relevant documents or passages given the query.  The Retrieval-Augmented Generation (RAG) model has achieved a significant improvement in the field of ODQA~\cite{lewis2020retrieval,abdallah2024arabicaqa,gruber2024complextempqa}. RAG models combine retrieved documents and advanced pre-trained large language models (LLMs) generating responses based on the retrieved information
~\cite{lewis2020retrieval,lala2023paperqa,mozafari-etal-2024-exploring,abdallah2023exploring,piryani-etal-2024-detecting}. However, the performance of RAG models depends on the top retrieved documents, especially on the first document~\cite{setty2024improving,zhang2024raft,abdallah2023generator}. The RAG model usually uses the first retrieved document, which is the primary source for generating the response. In RAG, queries and documents are embedded in a shared representation space to enable efficient search before using a task-specific model to perform a deeper, token-level document analysis. 

In this paper, we introduce \model, a simple, effective, fast, and cost-efficient re-ranking method that leverages the concept of \textit{answer scent} which is analogous to the way in which animals track the scent of their food or prey~\cite{maxwell2018information}.  Cognitive psychologists~\cite{winerman2012tracking} have found that people search for information online in much the same way as animals hunt for food, leading to the establishment of the concept of \textit{information scent} in the Information Retrieval field. It refers to the trail of relevant information that leads a user to the correct answer. Our proposal is built upon a similar concept of tracing the answer scent.
It first utilizes larger LLMs like GPT-3.5 or Llama 3-70B to generate an answer scent. This is done just once, hence is computationally efficient. Subsequently, a smaller model such as T5 is employed to re-rank the documents based on the received answer scent. This two-tiered approach allows leveraging the generative capabilities of a larger LLM to boost the re-ranking capabilities of smaller models thanks to improved contextual understanding. Our method scores retrieved documents using a zero-shot answer scent, which relies on a pre-trained LLM to compute the likelihood of the document-derived answers aligning with the answer scent. This approach allows to rank documents not just based on their initial retrieval scores but also on the likelihood of containing an answer (via answer scent) and the degree to which they contain information that aligns with the expected answer. By applying a cross-attention mechanism to every token in both the question and the passage, \model tracks the answer scent within the document corpus. Our approach successfully addresses the challenge of ensuring that the most relevant document is ranked at the top, which is a key component in open-domain question answering and RAG systems (Figure \ref{fig:intro}).

%% file: sections/methods.tex
\section{Method}
\begin{figure*}[t!]
\centering
\includegraphics[width=\textwidth,height=0.26\linewidth]{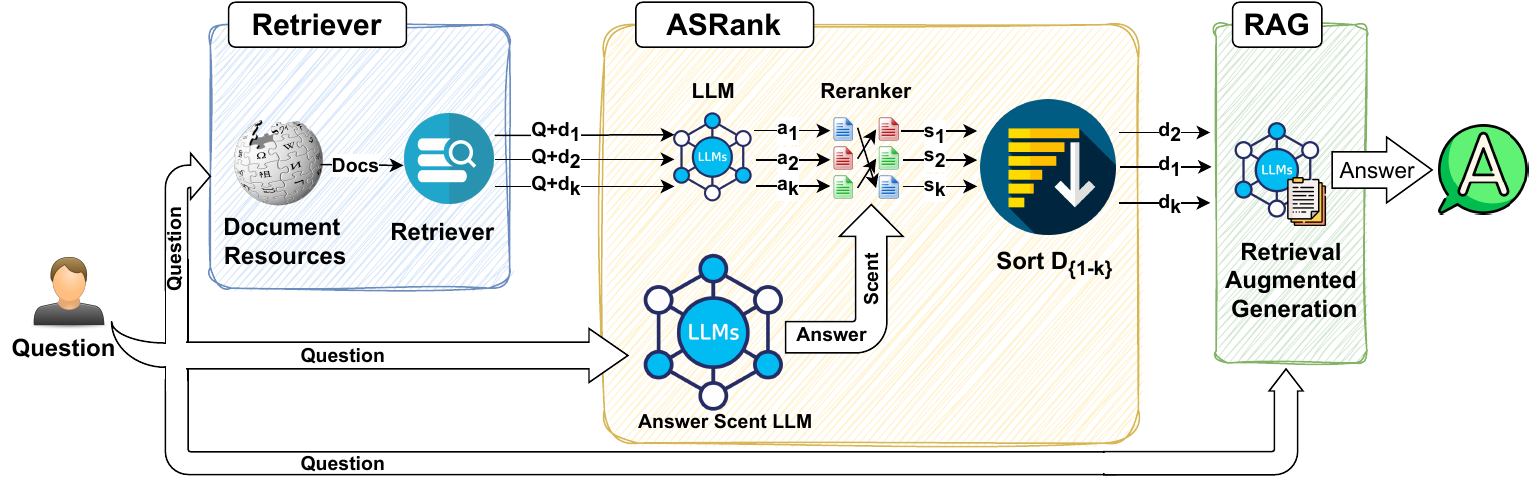}
\caption{
Our \model framework, starts with document retrieval, re-ranking using the answer scent from LLMs, and finally passing the top-k document into the RAG system.
}
\label{fig:framework}
\end{figure*}

In this section, we detail the methodology of \model, starting with retrieving documents based on either sparse or dense techniques. Subsequently, we introduce our concept of generating an Answer Scent using a large language model (Section \ref{sec:in-context-answer-scent-generation}), followed by an efficient re-ranking process that employs a smaller model (Section \ref{sec:zerorank-re-ranking}), which enhances the alignment and relevance of the retrieved documents to the query in our RAG system. Figure \ref{fig:framework} shows an overview of the \model framework.
\label{sec:method}

\subsection{Retriever}
\label{sec:method-retriever}

Let $\mathcal{M} = \{\boldsymbol{d}_1, \ldots, \boldsymbol{d}_M \}$ represent a collection of evidence documents. Given a query $\boldsymbol{q}$, the retriever's task is to select a subset of relevant documents $\mathcal{D} \subset \mathcal{M}$, aiming to include those that likely contain the answer to $\boldsymbol{q}$. Our framework is designed to operate on documents retrieved by arbitrary methods, hence ones that can either utilize sparse or dense representations. \textbf{Sparse representation} methods such as BM25~\cite{robertson2009probabilistic}, a non-neural approach, typically rely on term frequency and inverse document frequency to rank documents. This method is effective for scenarios where lexical matching is crucial, providing a strong baseline due to its simplicity and proven efficiency in various information retrieval tasks. \textbf{Dense representation} methods like Dense Passage Retrieval (DPR)~\cite{karpukhin2020dense} employs neural network architectures to encode queries and documents into dense vector spaces. The relevance of documents is assessed based on the similarity of these vectors, allowing to capture of semantic relationships that go beyond keyword matching. Regardless of the retrieval technique employed, the retrieval system identifies the top-K most relevant documents, denoted as $\mathcal{D} = \{\boldsymbol{d}_1, \ldots, \boldsymbol{d}_K\}$. 

\subsection{Answer Scent Generation}
\label{sec:in-context-answer-scent-generation}

Large language models (LLMs) such as GPT-3~\cite{brown2020language} and LLaMA~\cite{llama_v2} define probability distributions over sequences of tokens. Given a sequence $x_1,...,x_n$, these models typically predict the sequence's probability using an autoregressive approach $
p(x_1,...,x_n) = \prod_{i=1}^n p(x_i|x_{<i}),
$ where $x_{<i} := x_1,...,x_{i-1}$ represents the sequence of tokens preceding $x_i$, also referred to as its \emph{prefix}. This modeling is implemented via a transformer network parameterized by $\theta_1$, typically employing a causal self-attention mask $
p(x_1,...,x_n) = \prod_{i=1}^n p_{\theta_1}(x_i|x_{<i}),
$
which effectively models the conditional probabilities of each token.

In our approach, we incorporate the concept of \emph{answer scent}, which guides the model in generating answers that are contextually appropriate for the query, inspired by the success of In-Context Learning~\cite{brown2020language,ram2023context,dong2022survey}. This context is derived using a zero-shot approach, where the model infers the scent without explicit prior training on such task: $
p(x_1,...,x_n) = \prod_{i=1}^n p(x_i|x_{<i}, \mathcal{S}(x_{<i})),
$ where $\mathcal{S}(x_{<i})$ denotes the inferred answer scent in the form of a generated text content. The objective of $\text{Scent}$ is to encode the essence of what the answer should represent, enriching the input provided to the LLM reranker.

\vspace{-2mm}

\subsection{\model Re-Ranking}
\label{sec:zerorank-re-ranking}

\model introduces an unsupervised re-ranking utilizing LLM to evaluate the relevance of documents based on $\mathcal{S}(q)$, which serves as the guiding context corresponding to the target query $q$.

The core of \model's method is the calculation of a relevance score for each document, leveraging both the content of the document and its alignment with the inferred answer scent. The score is formulated as:
\[
{\small
\begin{aligned}
s(\boldsymbol{d}_i) = \sum_{t=1}^{|\boldsymbol{a}|} &-\log p(a_t \mid \boldsymbol{a}_{<t}, \boldsymbol{d}_i, \boldsymbol{q}, \mathcal{S}(\boldsymbol{q}); \theta_2),
\end{aligned}
}
\]

where $\boldsymbol{d}_i$ represents an individual document within the set of retrieved documents $\mathcal{D}$, $|\boldsymbol{a}|$ denotes the length of the $\boldsymbol{a}$ - an answer generated based on $\boldsymbol{d}_i$ by the rank model, and $\mathcal{S}(\boldsymbol{q})$ represents the answer scent derived from the query $\boldsymbol{q}$. The term $\log p(a_t \mid \boldsymbol{a}_{<t}, \boldsymbol{d}_i, \boldsymbol{q}, \mathcal{S}(\boldsymbol{q}); \theta_2)$ is the log probability of each token $a_t$ of the answer conditional on the prior tokens $\boldsymbol{a}_{<t}$, the document $\boldsymbol{d}_i$, the query $\boldsymbol{q}$, and the answer scent, parameterized by the model's parameters $\theta_2$.

To elaborate, the computation of the conditional probabilities can be decomposed as follows:

\[
{\small
\begin{aligned}
\log p(\boldsymbol{a} \mid \boldsymbol{d}_i, \boldsymbol{q}, \mathcal{S}(\boldsymbol{q})) = \sum_{t=1}^{|\boldsymbol{a}|} &\log p(a_t \mid \boldsymbol{a}_{<t}, \boldsymbol{d}_i, \boldsymbol{q}, \mathcal{S}(\boldsymbol{q}); \theta_2),
\end{aligned}
}
\]

where \(\log p(\boldsymbol{a} \mid \boldsymbol{d}_i, \boldsymbol{q}, \mathcal{S}(\boldsymbol{q}))\) represents the log probability of generating the answer \(\boldsymbol{a}\) given the document \(\boldsymbol{d}_i\), the query \(\boldsymbol{q}\), and the inferred answer scent \(\mathcal{S}(\boldsymbol{q})\).
The relevance score, denoted by \( s(\boldsymbol{d}_i) \), is reformulated using Bayes' Theorem. 
This score is represented as follows:
\[
{\small
\begin{aligned}
s(\boldsymbol{d}_i) \propto \log p(\boldsymbol{a} \mid \boldsymbol{d}_i, \boldsymbol{q}, \mathcal{S}(\boldsymbol{q})) 
+ \log p(\boldsymbol{d}_i \mid \boldsymbol{q}) - \log p(\boldsymbol{a} \mid \boldsymbol{q}),
\end{aligned}
}
\]
where  \(\log p(\boldsymbol{d}_i \mid \boldsymbol{q})\) is the log probability that the document \(\boldsymbol{d}_i\) is relevant to the query \(\boldsymbol{q}\), based on the initial retrieval. The normalization term \(-\log p(\boldsymbol{a} \mid \boldsymbol{q})\) adjusts for the base likelihood of the answer \(\boldsymbol{a}\) being related to the query \(\boldsymbol{q}\) across all documents. 

The decision to select the most relevant document employs a maximization approach $
\hat{i} = \arg\max_{i \in [1, K]} s(\boldsymbol{d}_i),
$, 
enhancing the likelihood that the document contains the information necessary to answer the query effectively, aligned with the derived answer scent.

%% file: sections/experimental_setup.tex
\section{Experiment Settings}
\vspace{-2mm}

\subsection{Datasets}
We utilize several common datasets for our experiments, whose detailed statistics are also provided in Appendix \ref{apx:dataset}:
\vspace{-0.2cm}
\paragraph{Open Domain QA datasets:}
TriviaQA~\cite{joshi2017triviaqa} is a collection of trivia questions sourced from trivia and quiz-league websites. 
Natural Questions (NQ)~\cite{Kwiatkowski2019natural} is a question-answering dataset containing 79,168 training examples, 8,757 development examples, and 3,610 test question-answer pairs. 
WebQuestions~\cite{berant-etal-2013-semantic} is a question-answering dataset that was created using Freebase as a knowledge base and which contains 5,810 question-answer pairs. 
\vspace{-0.2cm}
\paragraph{Entity-centric Questions:} EntityQuestions~\cite{sciavolino2021simple} contains 22K short questions about named entities based on facts from Wikipedia. 
\vspace{-0.2cm}
\paragraph{Temporal Questions:} ArchivalQA~\cite{wang2022archivalqa} is a large-scale question answer collection designed specifically for temporal news QA, containing 532,444 question-answer pairs, often on detailed or minor aspects. These pairs are derived from the New York Times Annotated Corpus, which spans news articles published between January 1, 1987, and June 19, 2007. We follow prior work~\cite{wallat2024temporal} and evaluate \model on the subset of ArchivalQA dataset, which comprises 7,500 questions. 


\paragraph{Multi-hop Questions:} HotpotQA~\cite{yang2018hotpotqa} 
contains 113K crowd-sourced questions constructed in a way that the introduction paragraphs of two Wikipedia articles are required to answer questions (i.e., two hops). We focus on the fullwiki setting, in which two Wikipedia passages are required to answer the questions. 
We follow prior work~\cite{khalifa2022few} and evaluate \model on the development set, which has 7,405 questions.
\paragraph{Information Retrieval:} BEIR is a diverse suite designed to test retrieval algorithms across multiple tasks, such as fact-checking and question-answering, with datasets from various domains including news, technical documents, and Wikipedia~\cite{thakur2021beir}. Specifically, we utilize the NFCorpus, DBPedia, Touche and News datasets from BEIR, which represent a range of retrieval challenges.  TREC-DL19~\cite{craswell2020overview} is a benchmark dataset widely used in IR research. We use the test sets of 2019 which contain 43 queries.

\subsection{Retrievers}  \label{subsec:retrievers_}
In our re-ranking experiments, we first retrieve passages using both unsupervised and supervised retrievers, as detailed below.

\paragraph{Unsupervised Retrievers:} BM25~\cite{Robertson2009bm25} is a ranking function used by search engines to estimate the relevance of documents to a given search query. Masked Salient Spans (MSS)~\cite{sachan2021end} is a dense retriever trained by predicting masked salient spans like named entities with the help of a reader network.  Contriever~\cite{izacard2021towards} is a framework for pre-training and fine-tuning models for information retrieval using contrastive learning.  

\paragraph{Supervised Retrievers:}

Dense Passage Retrieval (DPR)~\cite{karpukhin2020dense} uses annotated question-context paragraphs and hard negative examples to train a supervised dense retriever. MSS-DPR~\cite{sachan2021end} further improves DPR performance by first pre-training the dense retriever using MSS followed by DPR-style supervised fine-tuning. 

\subsection{LLM Models}
This section overviews the large language models (LLMs) utilized in our experiments. These models are essential for generating the "answer scent" and for re-ranking documents based on their inferred relevance to the query. 
\vspace{-0.2cm}
\paragraph{Answer Scent Models:}
We leverage a variety of Large Language Models (LLMs) to generate the answer sent, each bringing unique strengths to our re-ranking method. The \textbf{Llama} models, developed by Meta, are known for their robust performance in dialogue applications, having undergone extensive pre-training and fine-tuning~\cite{llama_v2}. \textbf{Mistral} and \textbf{Mixtral}, from Mistral AI, push the boundaries of efficiency and computational optimization, employing instruction fine-tuning and a sparse mixture of experts approach, respectively~\cite{mistral,mixtral}. \textbf{Gemma}, a product of Google, offers both base and instruction-tuned versions in different sizes, designed for adaptability across various hardware platforms~\cite{team2024gemma}. \textbf{GPT}, from OpenAI, is renowned for its general-purpose capabilities, pre-trained on vast data pools to generate semantically rich responses~\cite{brown2020language}. Lastly, \textbf{Qwen}, by Alibaba Cloud, encapsulates a broad pre-training regime across multiple languages and domains, optimized for long-context interactions, highlighting its scalability and depth in handling complex linguistic tasks~\cite{bai2023qwen}. 

\paragraph{Rank Model: }
In our experiments, we specifically utilize the T5 Base and T5 Large models, a variation of the original T5 architecture~\cite{raffel2020exploring} adapted for language modelling tasks. This architecture
features encoder and decoder transformers pre-trained to improve their ability to handle input text sequences.

\input{tables/tab-open-domain-rank}

\subsection{Experimental Setup}

\label{sec:experiment-setup}

All re-ranking experiments were conducted on a high-performance computing cluster using NVIDIA A100 48GB GPUs, while some experiments, such as ones in Section  \ref{result:rag} were done using NVIDIA A40 GPUs. We evaluated our method across five retrievers: BM25, MSS, MSS-DPR, DPR, and Contriever, retrieving the top 1,000 passages for re-ranking, consistent with the setup in~\citet{sachan2022improving}. For temporal questions in the ArchivalQA dataset, we also included Ance~\cite{xiong2020approximate} and RocketQA~\cite{qu2020rocketqa} for a comprehensive evaluation. Retrieval settings for the HotpotQA dataset followed the configurations from ~\citet{khalifa2022few} to ensure consistency. The UPR baseline was primarily evaluated using the T0-3B model, except in Table \ref{tab-qa-nq-reranking-plms}, where T0-3B, T0-11B, and T5-11B were used to explore model size variations. For BEIR and TREC datasets, we adopted the same settings as RankGPT~\cite{sun2023chatgptgoodsearchinvestigating}, using BM25 with 100 retrieved documents. To generate the answer scent, we employed the Llama-3-70B (instruction version) and GPT-3.5-turbo-0125 models, which were set to a temperature of 0.7 and a max length of 128 tokens. For re-ranking, T5-base and T5-large models were used, with a batch size of 128. To evaluate \model's performance, we used Top-K retrieval accuracy, following the methodology outlined in ~\citet{sachan2022improving}. For RAG evaluations, we measured performance using exact match, recall, and F1 scores. Additional details on the framework implementation and metrics can be found in Appendix \ref{sec:metrics} and \ref{ref:framework_imp}.

%% file: tables/tab-open-domain-rank.tex
\begin{table*}[t!]
\small
\centering
\centering
\begin{adjustbox}{width=0.75\textwidth}
\begin{tabular}{@{}l | c c c c | c c c c | c c c c  @{}} 
\toprule
\tf{Retriever} & \multicolumn{4}{c}{\tf{NQ }} & \multicolumn{4}{c}{\tf{TriviaQA}} & \multicolumn{4}{c}{\tf{WebQ}}  \\ 
                    & Top-1 & Top-5 & Top-10 & Avg & Top-1 & Top-5 & Top-10 & Avg & Top-1 & Top-5 & Top-10 & Avg \\
\midrule
\multicolumn{13}{c}{\textit{Unsupervised Retrievers}} \\
\midrule
MSS               & 19.2 & 41.2 & 51.2 &  37.2 & 30.7 & 52.6  & 60.5   & 47.9 & 11.6  & 29.0 & 39.1 & 26.6  \\ 
MSS + UPR$_{T0\_3B}$         &  38.7 & \textbf{64.8} & \textbf{72.2} &58.6 & 57.2 & 75.5  &  78.9 & 70.5 &  29.9 &  57.4 & 65.0 &  50.7 \\ 

MSS + \model$\dagger$$+T5_{base}$       &   45.2 & 64.7  & 70.6 & 60.1 &  65.3 &77.2  & 79.8  & 74.1 & 42.5  &61.3  & 67.7 &  57.1 \\
MSS + \model  $ \ddag  $$+T5_{base}$        &  \textbf{46.5} &  64.4 & 69.8 & \textbf{60.2} &  \textbf{66.3} & \textbf{77.6}  & \textbf{80.1}   & \textbf{74.6} &  \textbf{45.0} & \textbf{63.6} & \textbf{68.8} & \textbf{59.1}  \\ 
\midrule

BM25              & 22.1 & 43.7 & 54.4 &40.1 &  46.3 & 66.2  & 71.7 & 61.4 & 18.8 & 41.8 & 52.1 & 37.6  \\ 
BM25 + UPR$_{T0\_3B}$         &  35.4 &  63.4 & 70.2& 56.3 & 55.7 & 76.5 & 80.2      & 70.8 & 30.1  &57.3  & 66.5 &  51.3\\ 
BM25 + RANKGPT  $ \ddag  $     & 43.4 & 62.3 & 68.0 & 57.9 &-&  -&- & -& 40.3  &  57.6 &64.1 &  54.0  \\ 

BM25 + \model$\dagger$$+T5_{base}$       &  46.2 & 65.3  & \textbf{72.3} & 61.2 & 67.2  &  \textbf{77.9} & \textbf{80.7}  & 75.2 & 44.8  & 63.7 & 68.7 &   \textbf{59.0}\\

BM25 + \model   $ \ddag  $$+T5_{base}$      & \textbf{47.3} &  \textbf{65.6} &  71.4 & \textbf{61.4} & \textbf{67.3} & \textbf{77.9} & \textbf{80.7} &  \textbf{75.3} & \textbf{45.4} & \textbf{62.9}  & \textbf{68.9} & \textbf{59.0}   \\
\midrule
Contriever        &  22.1 & 47.2 & 58.7 & 42.7 & 34.1 &  59.4 &  68.0 & 53.8 &  19.9 &  43.4 & 56.3 & 39.9  \\ 
Contriever + UPR$_{T0\_3B}$   & 36.4 & 64.6  & 72.4 & 57.8&  56.7  & 77.0  & 80.2 & 71.3 &   30.0 &58.5  & 68.2 &  52.2 \\

Contriever + RANKGPT$\ddag$     &  44.4 & 64.8 &69.6 & 59.6  &- & -& -& -& 43.6 & 63.6 & 70.0 & 59.0  \\ 

Contriever + \model$\dagger$$+T5_{base}$        &  41.5  & 61.3  & 68.4 & 57.0 & 57.9 & 72.8 &  76.8  & 69.1 & 42.9 & 62.7 & 69.8 & 58.4 \\

Contriever + \model$ \ddag  $$+T5_{base}$   &  \textbf{48.0} & \textbf{66.6} &  \textbf{72.5} & \textbf{62.3}& \textbf{66.8} & \textbf{78.9}  & \textbf{81.4}   &  \textbf{76.0} &  \textbf{46.8} & \textbf{64.8} & \textbf{70.8} & \textbf{60.8}  \\ 
\midrule
\multicolumn{13}{c}{\textit{Supervised Retrievers}} \\
\midrule
DPR               &  48.6 & 68.7 & 74.5 & 63.9 & 57.4 &  72.4 &  76.5  & 68.7 & 44.8  & 65.0 & 70.6 & 60.1 \\ 
DPR + UPR$_{T0\_3B}$          &  42.5 & 70.6 & \textbf{78.1} & 63.8 & 61.3 & 78.7  &  81.9  & 74.0 &  34.9 &  63.6 & 71.7 & 56.7  \\ 
DPR + RANKGPT $\ddag$     & 48.6 & 68.7 & 74.5 &  63.9 & - & - & - & - & 44.8  &65.0 & 70.6 &  60.1 \\ 

DPR + \model$\dagger$$+T5_{base}$          &  50.2 & 69.9  &76.1 & 65.3 & 68.8 & 79.8  &  \textbf{82.4} &  77.0& 48.2  & \textbf{68.1} & 73.2 &  63.1 \\
DPR + \model  $ \ddag  $$+T5_{base}$        &\textbf{ 51.3} & \textbf{70.6} & 76.0 & \textbf{65.9} & \textbf{69.9} & \textbf{79.8}  & 82.1   & \textbf{77.3} & \textbf{ 49.3} & 67.3 & \textbf{73.4} & \textbf{63.3}  \\ 
\midrule
MSS-DPR           & 50.1 & \textbf{71.8} & \textbf{77.4} & \textbf{66.5} & 61.6 &  75.2 & 79.1 & 71.9 &  44.2 & 65.0 & 71.6 & 60.3  \\ 
MSS-DPR + UPR$_{T0\_3B}$      & 41.4 & 69.8 & 77.9 & 63.0 & 60.5 & 78.9 & 82.5 & 74.0 & 31.8  & 61.6 & 70.3 & 54.5  \\

MSS-DPR + \model$\dagger$$+T5_{base}$          &  48.8 & 69.3  & 76.1 &  64.7&  69.4 & 80.4  &  82.9 & 77.5 &  47.7 &\textbf{ 67.0 }&\textbf{ 73.0} &  62.5 \\
MSS-DPR  + \model $\ddag$$+T5_{base}$    &   \textbf{50.6} & 69.3 & 75.2 & 65.0 &  \textbf{69.9} &  \textbf{80.5} &  \textbf{82.9}  & \textbf{77.7} & \textbf{49.7}  & 66.6 & 72.6 & \textbf{62.9}  \\ 
\bottomrule
\end{tabular}
\end{adjustbox}

\caption{Top-1, 5, 10 retrieval accuracy of re-ranking methods including \model and baseline models on the NQ, TriviaQA and WebQ Datasets. $\dagger$ refers to Llama-3-70B, $\ddag$ refers to GPT-3.5-turbo-0125. For a comparison between LLama-2 7b vs UPR see Table \ref{tab-open-domain-rank-llama7b} in Appendix \ref{appendix:llama_upr}. Due to the Computational cost of RankGPT, we experimented with three retrievers (BM25, contriever, DPR) on two datasets (NQ, WebQA). 
}
\label{tab-open-domain-rank}
\end{table*}

%% file: sections/experimental_result.tex
\vspace{-2mm}

\section{Experiment Results}
\vspace{-2mm}

\label{sec:experiment-results}
In this section, we evaluate \model on a variety of question-answering tasks, leveraging several datasets to assess its performance. The datasets employed cover different QA challenges, ranging from open domain to entity-centric, temporal, BEIR, TREC and multi-hop questions. The primary objective is to evaluate \model's capability to rank the Top-{1, 5, 10} retrieved passages. For this purpose, an initial retrieval of 1,000 passages per question is conducted for reranking. On benchmarks, we compare \model with LLama 70B and GPT3.5-Turbo
with state-of-the-art supervised and unsupervised passage re-ranking methods. The baselines include UPR~\cite{sachan2022improving}, RankGPT~\cite{sun2023chatgptgoodsearchinvestigating}, HYDE~\cite{gao2022precise}, PathRetreiver~\cite{Asai2019}, MDR~\cite{mdr2021}, DrKit~\cite{drkit2020}, and PromptRank~\cite{khalifa2022few}. 

\vspace{-2mm}

\subsection{ODQA Re-ranking}
In this section, we focus on evaluating \model across several ODQA datasets (NQ, TriviaQA, and WebQ). Table \ref{tab-open-domain-rank} shows improvements in retrieval Top-K accuracy. \model enhances the retrieval of Top-K results across various settings, often outperforming the UPR model. For instance, when combined with the MSS retriever on NQ dataset, \model$\dagger$ increases the relevance of the Top-1 result to 45.2\%, a notable improvement over UPR's increase to 38.7\%. Similarly, TriviaQA dataset, \model with BM25 achieves a Top-1 accuracy of 47.3\%, surpassing BM25 + UPR's performance of 35.4\%. Also, the combination of \model with the MSS retriever results in a remarkable uplift in Top-1 accuracy for NQ, from an initial 19.2\% to 46.5\%. Similarly, on TriviaQA, \model with the BM25 retriever increases the accuracy of Top-1 from 22.1\% to 47.3\%.   In Appendix~\ref{apendix:case_study}, we show random examples from NQ dev and WebQA after and before re-ranking.  Further analysis of the effect of answer scent lengths is presented in Section~\ref{appendix:answer-scent-length}. The performance improvements across various LLMs are detailed in Section~\ref{appendix:Impact-of-Answer-Scent-LLM}

Additionally, we compare \model with RankGPT in Table \ref{tab-open-domain-rank}. \model outperforms RankGPT in retrieval accuracy across the NQ and WebQ datasets. For instance, \model$\ddag$ (i.e., \model with Llama-3-70B) achieves a Top-1 accuracy of 47.3\% on NQ, compared to RankGPT's 43.4\% with the BM25 retriever. In WebQ, \model$\ddag$ reaches a Top-1 accuracy of 45.4\%, surpassing RankGPT's 40.3\%. \textbf{Cost: }Utilizing RankGPT with GPT-3.5 across NQ and WebQ datasets, with DPR, Contriever, and BM25 retrievers, incurs a total cost of \$700. In comparison, running the same experiments with \model costs only \$15, demonstrating its cost-effectiveness at scale. Further details on performance improvements can be found in  Appendix~\ref{apendix:performance-improvement} while the impact of answer scent on re-ranking and latency implications are reported in Section~\ref{appendix:impact-answer-scent-Latency}.
\subsection{EntityQuestions Result}
\label{appendix:entity-result}
This subsection focuses on the evaluation of the Entity Questions dataset using various retrievers. The baseline models include MSS, DPR, MSS-DPR, and Contriever, and their retrieval performance is measured in terms of Top-1, Top-5, and Top-10 accuracy, as well as average accuracy across these metrics.  After applying re-ranking with Llama-3-70B, all retrievers see notable improvements in accuracy. Contriever, when combined with T5-base, achieves the highest Top-1, Top-5, and Top-10 scores, outperforming the baseline results by a large margin. The other retrievers, including MSS-DPR and DPR, also see significant gains after re-ranking, underscoring the effectiveness of using Llama-3-70B for enhancing retrieval performance.

The re-ranking with GPT-3.5-turbo-0125 produces similar improvements. Contriever again leads the performance metrics, achieving the highest Top-1 accuracy (48.9\%) and consistently strong Top-5 and Top-10 results. MSS-DPR also demonstrates robust improvements, further validating the ability of the re-ranking process to enhance retrieval accuracy for the Entity Questions dataset. 

\input{tables/tab-entity-questions}

\input{tables/tab-beir}
\vspace{-2mm}

\subsection{BEIR and TREC Results}

In this section, we evaluate the performance of \model on both the TREC (DL 19) and BEIR datasets, using nDCG@10 as the evaluation metric. For each dataset, the retriever selected the top 100 documents using BM25. Table \ref{tab-Comparsion-rankgpt-beir} presents the results from the DL 19 and BEIR datasets, which includes NFCorpus, DBPedia, and News. Across the datasets, \model method, using GPT-3.5 combined with T5 base, outperformed the other approaches, achieving the highest average score of 48.58. Specifically, \model achieved a score of 74.42 on the DL 19 dataset, surpassing both RankGPT and HyDE. While RankGPT performed slightly better on the News dataset, reaching 48.85 compared to \model’s 48.45, \model remained competitive across all datasets. The Llama 70B + T5 base variant of \model also delivered strong results, particularly on the DL 19 dataset, scoring 72.15.

\section{Additional Studies}
\subsection{Evaluation on NQ development}
In this section, we compare our approach with UPR using different model sizes (T0-3B, T0-11B)~\cite{sanh2021multitask} to assess the efficacy in the context of the NQ development set. This comparison highlights the significant advantages offered by the \model, across different retrievers like BM25, MSS, and DPR. The results are shown in Table \ref{tab-qa-nq-reranking-plms}. \model enhances retrieval performance across all Top-1, 5, 10, 20 results. Notably, after re-ranking with \model using the Llama 70B configuration, the accuracy of Top-1 for the MSS-DPR combination reaches 48.1\%, which is an improvement over its performance with UPR, where the Top-1 achieves 39.7\%. Similarly, the Top-1 for DPR alone ascends to 50.4\% with GPT3.5, surpassing the 41.1\% recorded with UPR.  Additionally, the performance of \model with T5-base + BM25 achieves a Top-1 accuracy of 47.5\%, surpassing previous results obtained with T5-lm-adapt (11B) or T0-11B.


\subsection{Evaluation on Diverse Question Answering Datasets}

The further studies were conducted across three distinct datasets—EntityQuestions, HotpotQA, and ArchivalQA. As summarized in Tables  \ref{tab-hotpot-questions}, \ref{tab-archival-questions}  (Appendix~\ref{appendix:archival-result}), and \ref{tab-entity-questions} (Section~\ref{appendix:entity-result}), \model enhances Top-1, 5, and 10 retrieval accuracies, across different retrievers.

\input{tables/tab-qa-nq-reranking-plms}
The EntityQuestions dataset, when used Llama 70B and GPT3.5 boosts performance, achieving Top-1 retrieval accuracy up to 48.9\%, which indicates an increase of over 25\% compared to baselines.  For the HotpotQA dataset, which requires reasoning over multiple documents, \model achieves substantial enhancements in Top-2, surpassing fully-supervised baselines like DPR~\cite{karpukhin-etal-2020-dense} and DrKit~\cite{drkit2020}, MDR~\cite{mdr2021}, PathRetriever~\cite{Asai2019} when combined with TF-IDF. This shows \model's strength in multi-hop question answering, supporting complex inference tasks across linked data points. Notably, \model combined with DPR achieves a Top-2 accuracy of 42.6\%, which not only surpasses the fully-supervised baselines such as DPR at 18.5\% and DrKit at 38.3\%, but also outperforms unsupervised models like PromptRank-GPT2-XL and PromptRank-T5-XL~\cite{khalifa2022few}, which score 36.6\% and 42.8\%, respectively. On the ArchivalQA dataset, which contains temporal questions, \model shows also good improvements. After re-ranking with Llama 70B and GPT3.5, the model significantly boosts Top-1 accuracies across different retrievers, demonstrating its effectiveness in extracting temporally relevant information. Specifically, after re-ranking with Llama 70B, BM25 improves from 18.2\% to 26.2\% in Top-1, DPR from 17.0\% to 27.5\%, and ANCE from 18.0\% to 27.3\%. When using GPT3.5, BM25 improves further to 27.6\% in Top-1, DPR reaches 27.7\%, and ANCE advances to 28.1\%.

\input{tables/tab-hotpot-questions}

\vspace{-0.3cm}

\subsection{Impact of Answer Scent on Re-Ranking and Latency Implications}
\label{appendix:impact-answer-scent-Latency}

Answer scents improve the alignment of retrieved documents with the question. The process is captured through the computation of log-likelihood of each document given the question $\log p(\boldsymbol{d} \mid \boldsymbol{q}; \Theta) = \frac{1}{|\boldsymbol{d}|}\sum_t \log p(d_t \mid \boldsymbol{d}_{<t}), $
where \( \boldsymbol{d} \) denotes the document tokens, \( \boldsymbol{q} \) the question, \( \Theta \) the LLM parameters, and \( |\boldsymbol{d}| \) the number of tokens in the document. 

\begin{figure}[h!]

\includegraphics[width=0.45\textwidth]{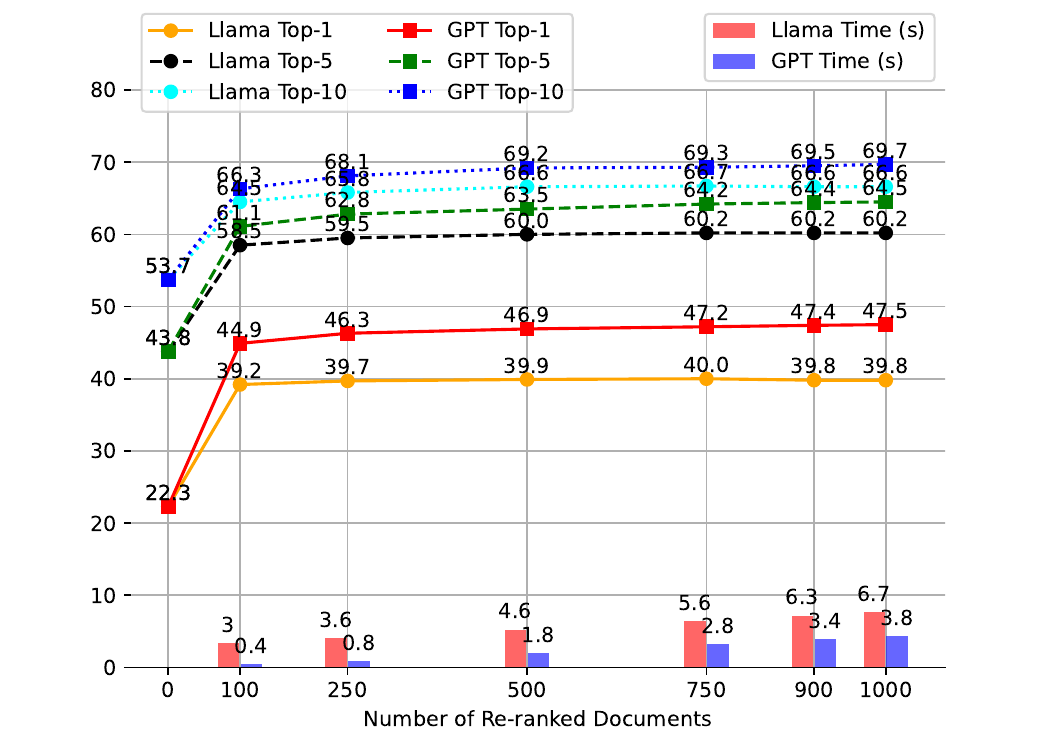}
\caption{
Effect of the number of passage candidates on the accuracy of Top-1, 5, 10 results, and latency when re-ranked with LLama 8B and GPT 175. The results were computed on the NQ development set using BM25 retrieved passages.
}
\label{fig:latency}
\end{figure}

Re-ranking with Answer Scent has shown improvements in Top-K on the NQ development set. The Top-1 accuracy increases from 22.3\% at 100 documents to around 39.8\% at 750 documents (see Fig. \ref{fig:latency}). The \model significantly reduces latency challenges, especially as the number of re-ranked documents grows. For example, re-ranking 1,000 documents takes up to 6.7 seconds with Llama models and 3.8 seconds with GPT models, compared to 11.6 seconds with UPR. This means \model cuts latency by nearly 42\% compared to UPR, as shown in Figure \ref{fig:latency}, and Table \ref{tab:qa-time} in Appendix \ref{ref:tab-time-sec}. 

\subsection{Role of Answer Scent}
\vspace{-0.1cm}

In this section, we present an ablation study evaluating the contribution of the answer scent in the \model framework. Table \ref{tab-no-answer} shows the results for different retrievers (DPR, MSS, and BM25) under various conditions: with an unknown (<UNK>) token used as the answer scent, and with the answer scent on NQ-test. 
For instance, DPR’s Top-1 accuracy improves from 23.4\% without answer scent to 51.3\% when the answer scent is provided. Similarly, BM25's Top-1 accuracy increases from 23.5\% to 47.3\%.
\begin{figure}[h!]
    \centering
    \includegraphics[width=0.85\linewidth]{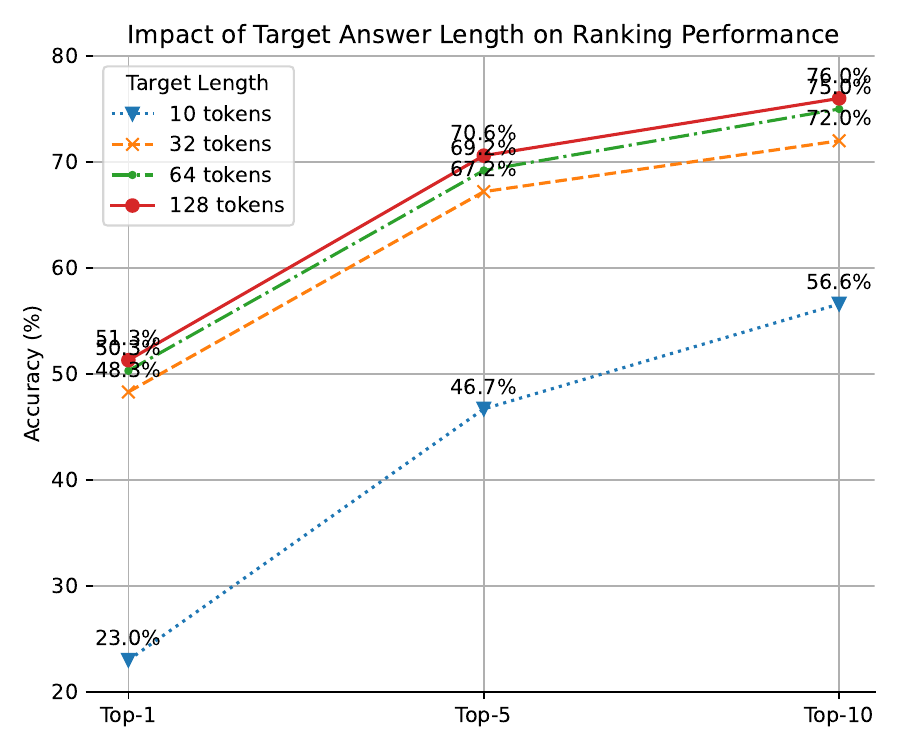}
    \caption{ Impact of target answer length on ranking performance using DPR on NQ-test.}
    \label{tab-answer-scent-length-2}
\end{figure}

\input{tables/tab-no-answer}
    
\begin{figure*}[t!]
\centering
\includegraphics[width=0.70\textwidth,height=0.27\linewidth]{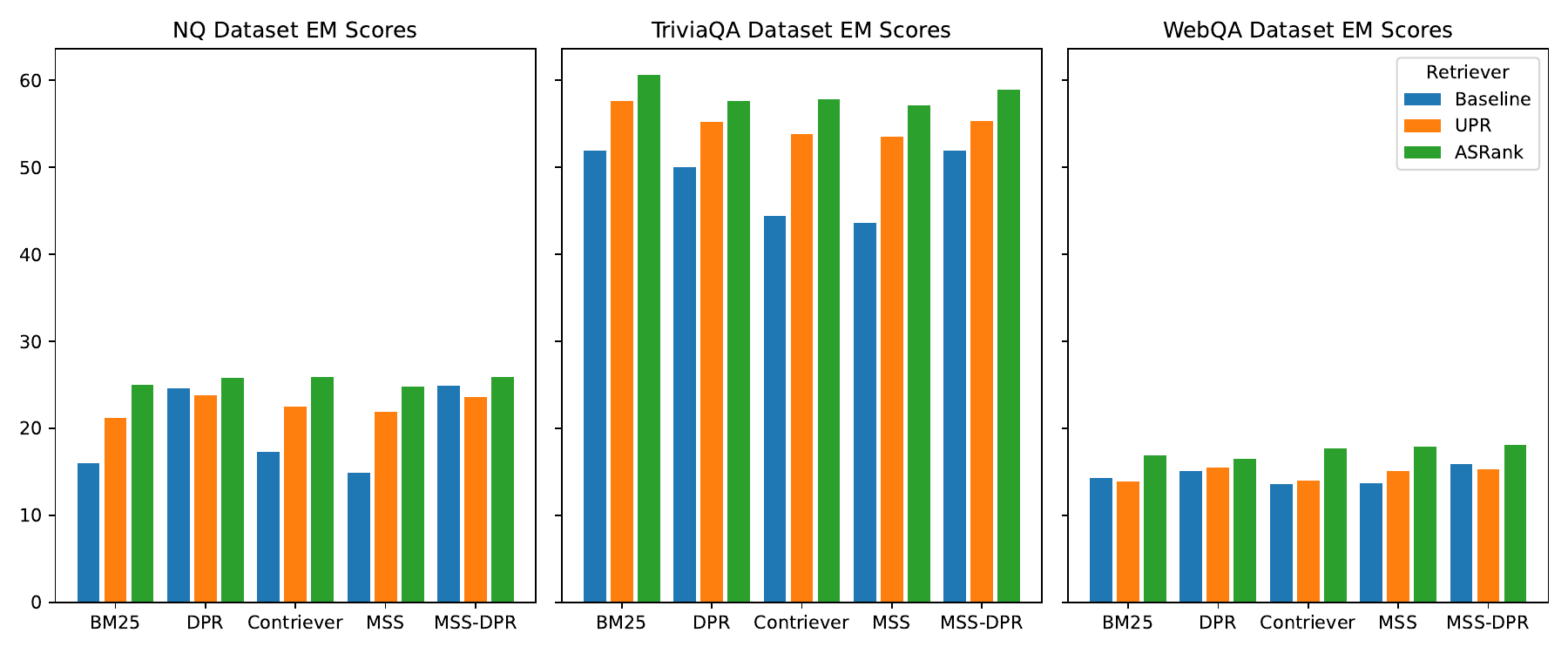}
\caption{Comparison of Exact Match (EM) scores across three datasets (NQ, TriviaQA, and WebQA) for various retrieval models.
}
\label{fig:em}
\end{figure*}

\subsection{Answer Scent Length}
\label{appendix:answer-scent-length}
This subsection analyzes the impact of varying Answer Scent (AS) lengths on retrieval performance using the DPR on the NQ-test dataset. Table \ref{tab-answer-scent-length} presents the results of four different AS lengths: 50, 75, 100, and 128 tokens, increasing the length of the answer scent improves retrieval performance. Starting with an AS length of 50 tokens, the model achieves a Top-1 accuracy of 42.6\%, Top-5 accuracy of 62.3\%, and Top-10 accuracy of 69.3\%, with an average accuracy of 58.0\%. When the AS length increases to 75 tokens, all metrics improve, particularly the Top-10 accuracy, which jumps to 78.2\%.

Further increasing the AS length to 100 and 128 tokens leads to even higher gains. With 128 tokens, the model achieves its best performance, reaching a Top-1 accuracy of 51.3\%, Top-5 accuracy of 70.6\%, and Top-10 accuracy of 76.0\%, resulting in an overall average of 65.9\%.

\input{tables/tab-answer-length}

\subsection{Impact of Answer Scent LLM}
\label{appendix:Impact-of-Answer-Scent-LLM}
In this section, we evaluate the impact of different LLMs on Top-{1, 5, 10, 20} by re-ranking the Top-1000 passages from the NQ development set. The performance of these LLMs on the NQ development set is detailed in Table \ref{tab-Different-large-llm}. The baseline retrieval using BM25 achieves a Top-1 of 22.3\%. However, with LLMs like Llama-2 and GPT3.5, there’s an increase in all Top-K. For instance, Llama-2 70B improves the Top-1 accuracy to 45.3\%, and GPT3.5 pushes it further to 46.3\%. As the model size increases from 7B to over 70B, there’s a performance improvement. The Mixtral model achieves a Top-1 of 42.5\%.   The increase in Top-K from Llama-2 70B to Qwen1.5 110B is marginal in Top-1 (from 45.3\% to 44.0\%) .This suggests that larger models have better re-ranking capabilities due to their ability to process a broader scope of linguistic nuances. however, while larger models generally perform better, the rate of improvement varies. For example, the increase in Top-K from Llama-2 70B to Qwen1.5 110B is marginal in Top-1 (from 45.3\% to 44.0\%), indicating diminishing returns at higher parameter counts. 
\input{tables/tab-Different-large-llm}

\subsection{Impact of Answer Length}
 We conducted two sets of experiments to further analyze the potential influence of answer length~\cite{wang2022perplexity} on ranking performance. The first experiment focused on the length of the generated answer scent, while the second examined the target length of answers provided by the T5 model. Table \ref{tab-answer-scent-length} presents the results for different lengths of the answer scent. We tested varying lengths (50, 75, 100, and 128 tokens) and found that 128 tokens consistently produced better results regarding Top-1, Top-5, and Top-10 accuracy, with DPR’s Top-1 accuracy reaching 51.3\%. Fig \ref{tab-answer-scent-length-2} explores the impact of different target answer lengths on ranking accuracy. The results show that increasing the target answer length generally improves the ranking, with the best performance observed at a target length of 128 tokens.

\vspace{-2mm}

\section{RAG for ODQA}

\label{result:rag}

\paragraph{Method}

In the Retrieval-Augmented Generation (RAG) framework, we employ a large language model (LLM), leveraging its capacity to utilize retrieved documents dynamically for generating responses. The RAG method combines the robust retrieval capabilities of DPR with the generative models, thereby enabling understanding and response generation based on the context provided by the retrieved documents. The RAG model is formulated as $ p(a \mid q, D) = \sum_{d \in D} p(d \mid q) \cdot p(a \mid q, d), $ where \(a\) represents the answer, \(q\) the query, and \(D\) is the set of retrieved documents relevant to \(q\). The term \(p(d \mid q)\) denotes the document's retrieval probability, and \(p(a \mid q, d)\) represents the probability of generating answer \(a\) given the query \(q\) and document \(d\).

\paragraph{Results}
We evaluated the RAG method on NQ, TriviaQA, and WebQA revealing significant performance gains as shown in Figure \ref{fig:em}. For instance, before applying our \model re-ranking strategy, the BM25+LLama7B achieves a baseline EM of 16.0\% on NQ. After re-ranking with \model, the EM increased to 24.8\%.  We show a detailed comparison between the baselines (BM25, DPR, MSS, Contriever, MSS-DPR), UPR, and \model in Table \ref{tab-rag} (Appendix \ref{appendix-rag}).

%% file: tables/tab-entity-questions.tex
\begin{table}[h!]
\small
\centering
\begin{adjustbox}{width={0.35\textwidth}}
\begin{tabular}{l | c c c c c }
 \toprule
 \tf{Retriever} & \multicolumn{4}{c}{\tf{Entity Questions}}  \\
               & Top-1& Top-5& Top-10& Avg \\ 
\midrule
\multicolumn{5}{c}{\textit{Baselines}} \\
\midrule
MSS  & 19.3 & 35.9 & 43.1&  32.8\\ 
DPR   & 25.3 & 39.5 & 45.3& 36.7 \\ 
MSS-DPR    & \textbf{30.3} & 47.7 & 54.1&  \textbf{44.0}\\ 
Contriever & 27.1 & \textbf{48.0}& \textbf{55.7}&  43.6\\ 
\midrule
\multicolumn{5}{c}{\textit{After Re-ranking with Llama-3-70B}} \\
\midrule
MSS$+T5_{base}$    & 44.5 &58.3 &  62.7 & 55.2\\ 
DPR$+T5_{base}$     & 41.7 & 53.8 & 58.2 & 51.2 \\ 
MSS-DPR$+T5_{base}$      & 46.4 & 60.1 & 64.5 & 57.0\\ 
Contriever$+T5_{base}$    & \textbf{46.6} & \textbf{61.1} & \textbf{65.9} &  \textbf{57.9}\\ 
\midrule
\multicolumn{5}{c}{\textit{After Re-ranking with GPT-3.5-turbo-0125}} \\
\midrule
MSS$+T5_{base}$   &  46.6 & 60.5 &  64.5 & 57.2\\ 
DPR$+T5_{base}$     & 43.6 & 55.6 & 59.4  & 52.9\\ 
MSS-DPR$+T5_{base}$      & 48.4 &  62.1 & 66.2  &58.9 \\ 
Contriever$+T5_{base}$  & \textbf{48.9} & \textbf{63.2} & \textbf{67.5}  & \textbf{59.8}\\

\bottomrule
\end{tabular}
\end{adjustbox}
\caption{Top-1, 5, 10 retrieval accuracy for the Entity Questions dataset, comparing baseline retrievers with results after re-ranking using Llama-3-70B and GPT-3.5-turbo-0125 models. 
}
\label{tab-entity-questions}
\end{table}

%% file: tables/tab-beir.tex
\begin{table}[ht!]
\centering
\begin{adjustbox}{width={0.5\textwidth}}
\begin{tabular}{@{}l | c | c c c c c | c @{}} 
\toprule
\tf{Method} &\tf{Model} & \multicolumn{1}{c}{\tf{DL 19}} & \multicolumn{1}{c}{\tf{NFCorpus}}& \multicolumn{1}{c}{\tf{Touche}} & \multicolumn{1}{c}{\tf{DBPedia}} & \multicolumn{1}{c|}{\tf{News}} & Avg \\ 
\midrule
BM25 & - & 50.58  & 30.75 & \textbf{44.22} & 31.80 & 39.52 & 39.37 \\
\midrule
UPR & $T5_{XL}$  &  53.85  & 35.04& 19.69 & 30.91   & 43.11 & 36.52 \\

monoBERT & BERT &  70.50  & 36.88&  31.75 &41.87  & 44.62 & 45.12 \\

monoT5 & $T5_{Base}$  &  71.48  & 37.38& 30.82 &42.42   & 46.83 & 45.78 \\

Cohere Rerank-v2 & - &  73.22  &  36.36 & 32.51 &42.51 &  47.59 & 46.43\\ 

Promptagator++ (few-shot)&  -  &  -&  37.0&  38.10 &  43.4 & - & - \\

RankGPT & GPT-3.5 &  65.80 & 35.62& 36.18 &44.47  &  \textbf{48.85} & 46.18 \\ 
HyDE    & InstructGPT &  61.30 &- & -&  36.80 &   44.00&  -\\

\model     &Llama-3-70B$+T5_{Base}$ & 72.15	 & 33.90 & 33.22 &	43.51 &  	41.13& 44.78 \\ 

 \model        & GPT-3.5$+T5_{Base}$& \textbf{74.42} & 	\textbf{38.15}& 35.56	&\textbf{45.37}	  & 48.45 & \textbf{48.39}\\ 
\hline
\end{tabular}
\end{adjustbox}
\caption{
Results (nDCG@10) on TREC and BEIR. GPT-3.5 is turbo-0125.
}
\label{tab-Comparsion-rankgpt-beir}
\vspace{-2mm}

\end{table}

%% file: tables/tab-qa-nq-reranking-plms.tex
\begin{table}[ht!]
\small
\centering
\begin{adjustbox}{width={0.45\textwidth}}
\begin{tabular}{@{}l | c c c c c @{}}
 \toprule
 \tf{Retriever } & \multicolumn{4}{c}{\tf{NQ (dev)}}  \\
                   &  Top-1 & Top-5 & Top-10  & Top-20 & Avg  \\
\midrule
MSS        & 17.7  & 38.6  & 48.7 & 57.4   & 40.6  \\
BM25       & 22.3  & 43.8  & 53.6 &62.3   &  45.5  \\  
Contriever & 19.6 & 45.4 & 55.8 & 64.9  &  46.4 \\
MSS+BM25      & 17.6  & 38.7 & 48.8 & 57.8 & 40.7  \\
MSS-DPR & \textbf{48.9} & \textbf{69.9} &  \textbf{75.7} & \textbf{80.4} &\textbf{ 68.7} \\
DPR & 47.8 & 67.3 &  73.0 & 77.4  &  6.4 \\
 \midrule
\multicolumn{5}{c}{\textit{After Re-ranking with UPR}} \\
\midrule
MSS+T0-3B  &36.6 & 62.9 &70.8 & 75.7 &  61.5 \\
BM25+MSS+T5-lm-adapt (3B) &  29.7  & 59.9& - & 76.9   & 55.5 \\
BM25+MSS+T5-lm-adapt (11B) &  32.1 & 62.3& -& 78.5 &  57.6 \\
BM25+MSS+ T0-3B            & 36.7 & 64.9& -& 79.1 &   60.2 \\
BM25+MSS+ T0-11B           & 37.4 & 64.9&- & 79.1 & 60.5\\
MSS+DPR+T0-3B  & 39.7 & 68.6 & 76.5 & \textbf{82.0 }& 66.7 \\
DPR +T0-3B & \textbf{41.1}  & \textbf{69.5} & \textbf{77.0}& 81.9 & \textbf{67.4} \\
\midrule
\multicolumn{5}{c}{\textit{After Re-ranking with Llama-3-8B}} \\
\midrule
BM25$+T5_{base}$   & 39.8 & 58.6 & 66.6 & 71.9&   59.6 \\
BM25+MSS$+T5_{Large}$ & \textbf{41.7} & \textbf{64.8} & \textbf{73.5} & \textbf{80.1}&  \textbf{65.0} \\
\midrule
\multicolumn{5}{c}{\textit{After Re-ranking with GPT-3.5-turbo-0125}} \\
\midrule
MSS$+T5_{base}$ & 46.2 & 63.5 & 69.1 & 73.2 & 63.0 \\
BM25$+T5_{base}$  &47.5 & 64.5 & 69.7 &74.3 &   64.0 \\
Contriever$+T5_{base}$  & 47.7 & 65.5 &71.2 & 76.2 & 65.2 \\
BM25+MSS$+T5_{base}$  &  47.9 & 65.5 & 71.2 & 76.4 &  65.3 \\
MSS-DPR$+T5_{base}$  & 50.1 & 68.9 & 74.8 & \textbf{79.8 }&  68.4\\
DPR$+T5_{base}$  & \textbf{50.4} &\textbf{68.9 }& \textbf{74.9} & 79.4 &  \textbf{68.4} \\
\midrule
\multicolumn{5}{c}{\textit{After Re-ranking with Llama-3-70B}} \\
\midrule
MSS$+T5_{base}$   &  44.9 & 63.7 & 69.4 & 73.9 &  62.9 \\
BM25$+T5_{base}$ & 44.8 & 64.1 &69.9 & 75.0 &   63.5 \\
Contriever$+T5_{base}$ & 45.7 & 65.4 & 71.5 & 76.2 &  64.7 \\
BM25+MSS$+T5_{base}$  &  45.4 &  65.4 & 71.0 & 76.6 & 64.6 \\
MSS-DPR$+T5_{base}$ & 48.1 & \textbf{68.6 }& \textbf{74.6} & \textbf{79.8} &\textbf{67.8}  \\
DPR$+T5_{base}$ & \textbf{48.2} & 67.9 & 73.8 & 78.6 & 67.1 \\

\bottomrule

\end{tabular}
\end{adjustbox}
\caption{Performance comparison of different retrievers on the NQ development set.
}

\label{tab-qa-nq-reranking-plms}
\vspace{-2mm}

\end{table}

%% file: tables/tab-hotpot-questions.tex
\begin{table}[h] 
\centering
\begin{adjustbox}{width={0.40\textwidth}}
\begin{tabular}{l|cccc}
\toprule

  \tf{Retriever}  & \multicolumn{4}{c}{\tf{HotpotQA}}  \\
  
  & \textbf{\# Ex.} &    \textbf{top-2} &  \textbf{top-10} &  \textbf{top-20} \\
  \midrule
\multicolumn{5}{c}{\textit{Fully-supervised Baselines}} \\
  \midrule
  DPR  & -  & 18.5  & 37.2 & 47.1 \\   
DPR+\model  & -  & 42.6 & 68.8& 79.2 \\
 DrKit  & \textasciitilde 90K            & 38.3  & 67.2  & 71.0  \\ 
 MDR  & \textasciitilde 90K             & 65.9  & 77.5  & \textbf{80.2}  \\
PathRetriever  & \textasciitilde 90K    & \textbf{66.4}  & \textbf{77.8}  & 78.7  \\

   \midrule
\multicolumn{5}{c}{\textit{Unsupervised Baselines}} \\
\midrule
 TF-IDF         & -- &  9.9  & 27.6  & 35.0   \\
 TF-IDF + BM25   & --& 19.1  & 54.7   & 61.8   \\
 
 PromptRank-GPT2-XL & -    & 36.6  & 60.5 & 65.9  \\
 PromptRank-T5-XL & -    & 42.8  & 68.9 & 74.1  \\

 TF-IDF+\model $\dagger$$+T5_{base}$  & -  &  36.9 &  61.1 &  72.5 \\
 TF-IDF+\model $\ddag$ $+T5_{base}$ & -  & \textbf{45.1 } & \textbf{69.1}  & \textbf{78.9}  \\

\bottomrule
\end{tabular}
\end{adjustbox}
\caption{
Top-2, 10, 20 retrieval performance on HotpotQA comparing \model to baselines. $\dagger$ refers to Llama-3-70B and $\ddag$ refers to GPT-3.5-turbo-0125.
}
\label{tab-hotpot-questions}
\end{table}

%% file: tables/tab-no-answer.tex
\begin{table}[h!]
\centering
\begin{adjustbox}{width={0.40\textwidth}}
\begin{tabular}{@{}l | c | c c c  @{}} 
\toprule
\tf{Retriever} &\tf{Mode} & \multicolumn{1}{c}{\tf{Top-1}} & \multicolumn{1}{c}{\tf{Top-5	}} & \multicolumn{1}{c}{\tf{Top-10}} \\

\midrule
\multirow{2}{*}{MSS}	&<UNK> &	22.9	&29.78&	46.9\\
	& Answer	&46.5	&64.4	&69.8\\
\midrule

\multirow{2}{*}{BM25}&	<UNK> &	23.5&	30.6	& 45.9\\
&	Answer&	47.3&	65.6&	71.4\\
\midrule
\multirow{2}{*}{DPR} & <UNK> 	 & 23.4	&31.9	&49.4       \\ 

	& Answer& 	51.3	& 70.6	& 76.0\\

\bottomrule

\end{tabular}
\end{adjustbox}
\caption{Performance Comparison with and without Answer Scent for NQ test dataset.}
\label{tab-no-answer}
\end{table}

%% file: tables/tab-answer-length.tex
\begin{table}[t!]
\centering
\begin{adjustbox}{width={0.4\textwidth}}
\begin{tabular}{@{}l | c | c c c c @{}} 
\toprule
\tf{Retriever} &\tf{AS length	} & \multicolumn{1}{c}{\tf{Top-1}} & \multicolumn{1}{c}{\tf{Top-5	}} & \multicolumn{1}{c}{\tf{Top-10}} & \multicolumn{1}{c}{\tf{Avg}} \\

\midrule
\multirow{4}{*}{ DPR}	 & 50	 & 	42.6	&62.3	&	69.3 & 58.0 \\ 

	& 75 & 	46.5	&	64.5 &	78.2 & 63.0 \\
	&100 &		47.5	&66.1		&	73.7 & 62.4\\
	& 128	&	51.3	& 70.6&76.0	 &	65.9
\\
			
\hline
\end{tabular}
\end{adjustbox}
\caption{ Impact of answer scent length on ranking performance using DPR on NQ-test.
}
\label{tab-answer-scent-length}
\end{table}

%% file: tables/tab-Different-large-llm.tex
\begin{table}[!ht]
\small
\centering
\begin{adjustbox}{width={0.45\textwidth}}
\begin{tabular}{@{}l | c c c c c c @{}}
 \toprule
 \tf{Retriever } & \tf{\#Parameters} & \multicolumn{5}{c}{\tf{NQ (dev)}}  \\
          &          &  \tf{Top-1} & \tf{Top-5} & \tf{Top-10} & \tf{Top-20} & \tf{Avg } \\
\midrule
       BM25   &  -  & 22.3  & 43.8  & 53.7 & 62.3   &    \\  
\midrule
Gemma   &  7B  & 21.2 & 37.7 & 45.9 & 54.2 &  39.8 \\
Mistral & 7B  & 27.9 & 46.3 &   54.8 & 62.3 &  47.8 \\
Qwen1.5 &  7B  & 30.3 & 50.4 & 58.6  & 66.2 & 51.4  \\
Llama-2 & 7B  & 39.2 &  58.6 & 65.8 & 71.4 &  58.8 \\
Llama3  & 8B  &  39.8  & 60.2 & 66.6 & 71.9 & 59.6  \\
Qwen1.5 &  14B  & 34.9  & 54.4 & 62.7 & 69.1 & 55.3  \\
Qwen1.5 &  32B  & 39.9 & 60.3 & 67.2 & 72.9 &  60.1 \\
Mixtral & 8x7B  &  42.5 & 61.9 & 68.2 &73.0 &  61.4 \\
Llama3  & 70B  & 44.8 & 64.1 & 69.9 & \textbf{75.0} & \textbf{63.5} \\
Qwen1.5 &  72B  & 43.2 & 62.6 & 68.9 & 73.9 &  62.2 \\
Llama-2 & 70B  & 45.3 &64.0  &\textbf{69.9 }& 74.4& 63.4 \\
Qwen1.5 &  110B  &  44.0 & 63.3 &  69.8 &  74.4 &  62.9 \\
GPT3.5  &  175B & \textbf{46.3} & \textbf{63.6} & 69.1 &  73.8 & 63.2\\ 
    
\bottomrule
\end{tabular}
\end{adjustbox}
\caption{
Performance metrics of different LLMs utilizing the answer scent concept for document retrieval across Top-1, 5, 10, and 20 rankings on the NQ (dev) dataset.
}
\label{tab-Different-large-llm}
\end{table}

%% file: sections/related_work.tex
\section{Related Work}



Recent developments in the field of information retrieval have increasingly focused on the integration of LLMs for enhancing retrieval and reranking mechanisms. LLMs have demonstrated a substantial impact in retrieval tasks, largely due to their deep generative capabilities. Innovative approaches like InPars~\cite{bonifacio2022inpars, jeronymo2023inpars} and Promptagator~\cite{dai2022promptagator} have explored the generation of synthetic datasets to improve domain-specific retrieval performance. Concurrently, models like SGPT~\cite{muennighoff2022sgpt} and UPR~\cite{sachan2022improving} have showcased the direct utility of GPT-based and T5 models as effective rankers in bi-encoder architectures, with UPR utilizing query likelihood for scoring.

Notably, PRP~\cite{qin2023large} and \citet{ma2023fine} have demonstrated that fine-tuning LLMs like LLaMA enhances retrieval performance beyond smaller models, positioning LLMs as powerful tools for reranking tasks. The integration of unsupervised and supervised retrieval techniques such as BM25~\cite{robertson2009probabilistic}, MSS~\cite{sachan2021end}, Contriever~\cite{izacard2021towards}, and DPR~\cite{karpukhin2020dense} has been pivotal. These methods, including enhancements like MSS-DPR~\cite{sachan2021endtoend}, leverage dense and sparse retrieval techniques to enhance the initial retrieval stages, subsequently improved through reranking. Moreover, newer supervised methods like ColBERT~\cite{khattab2020colbert} and SPLADE~\cite{formal2021splade} further refine retrieval accuracy. A growing body of work has investigated the role of LLMs in reranking by prompting them to reorder documents based on relevance, with methods like RankVicuna~\cite{pradeep2023rankvicuna} and LRL~\cite{ma2023zero} demonstrating significant advancements. These studies illustrate that LLMs with prompts can handle reranking tasks efficiently.

%% file: sections/Appendix.tex
\appendix
\section{Datasets' Details} \label{apx:dataset}

In this section, we present a table that details the statistics of the datasets utilized in our study. These tables include comprehensive data such as sample sizes, feature counts, and other relevant metrics, providing a clear overview of the datasets' composition and scope. 

\begin{table}[htb]
          \small
	\center
	\caption{Statistics of datasets.}
	\label{tbl:dataset_statistics}
	\begin{tabular}{@{}l|lll@{}}
		\toprule Dataset  & Train  & Dev   & Test   \\
		\midrule TriviaQA & 78,785 & 8,837 & 11,313 \\
		NQ                & 79,168 & 8,757 & 3,610  \\
		WebQA              & 3,417  & 361   & 2,032  \\
        Entity Questions   & - &  -  &  22,000 \\
        HotpotQA   & 90,564 &   7,405  &  - \\
        ArchivalQA   & - &   - &   7,500\\
		\bottomrule
	\end{tabular}
\end{table}

\section{Evaluation Metrics}
\label{sec:metrics}

To assess the performance of the \model, we use top-K retrieval accuracy and several other metrics for the RAG. Top-K retrieval accuracy measures whether the correct answer appears within the top-K retrieved passages, calculated as:
\[
\text{TOP@}k = \frac{\sum(\text{any(Correct@}k\text{))}}{\text{Total Data}}
\]

Given that LLMs tend to generate verbose answers, many standard QA metrics are not well suited to evaluate the answer quality; the Exact Match will always be less given the occurrence of other non-ground-truth tokens, and the F1 score will be penalized by other, potentially helpful tokens. Therefore, we utilize a set of model-agnostic metrics (i.e., token recall and answer string containment).

\section{Retrievers}  \label{apendix:retrievers}
In our re-ranking experiments, we retrieve passages using both unsupervised and supervised retrievers, as detailed below.

\paragraph{Unsupervised Retrievers} BM25~\cite{Robertson2009bm25} is a ranking function used by search engines to estimate the relevance of documents to a given search query. It is based on the probabilistic retrieval framework and uses term-frequency (TF) and inverse document frequency (IDF) of the keywords present in the question and passage. 

Masked Salient Spans (MSS)~\cite{sachan2021end} is a dense retriever trained by predicting masked salient spans like named entities with the help of a reader network. The objective function for training the MSS retriever can be represented as:

\begin{align*}
\mathcal{L}_{MSS} = -\mathbb{E}_{(q,d^+,d^-) \sim D}[\log p(d^+|q) \\ + \log(1 - p(d^-|q))]
\end{align*}

where $D$ is the dataset, $(q,d^+,d^-)$ is a triplet of the question, positive document, and negative document, and $p(d|q)$ is the probability of a document $d$ being relevant to a question $q$.

Contriever is a framework for pre-training and fine-tuning models for information retrieval using contrastive learning. The objective function for training the Contriever model is:

\begin{align*}
\mathcal{L}_{Contriever} = -\mathbb{E}_{(q,d^+,d^-) \sim D}[\log \sigma(s(q,d^+)) \\ + \log(1 - \sigma(s(q,d^-)))]
\end{align*}

where $s(q,d)$ is the similarity score between question $q$ and document $d$, and $\sigma$ is the sigmoid function~\cite{izacard2021towards}.

\paragraph{Supervised Retrievers}

 Dense Passage Retrieval (DPR)~\cite{karpukhin2020dense} uses annotated question-context paragraphs and hard negative examples to train a supervised dense retriever. The objective function for training the DPR model is:

\begin{align*}
\mathcal{L}_{DPR} = -\mathbb{E}_{(q,d^+,d^-) \sim D}[\log \sigma(s(q,d^+)) \\ + \log(1 - \sigma(s(q,d^-)))]
\end{align*}

where $s(q,d)$ is the similarity score between question $q$ and document $d$, and $\sigma$ is the sigmoid function.

MSS-DPR~\cite{sachan2021end} is an approach that further improves DPR performance by first pre-training the dense retriever using MSS followed by DPR-style supervised fine-tuning. The objective function for training the MSS-DPR model is:

\begin{align*}
\mathcal{L}_{MSS-DPR} = \alpha \mathcal{L}_{MSS} + (1 - \alpha) \mathcal{L}_{DPR}
\end{align*}

where $\alpha$ is a hyperparameter that controls the trade-off between the MSS and DPR losses.

\section{Implementation Framework}
\label{ref:framework_imp}
Our implementation of \model utilizes the PyTorch~\cite{paszke2019pytorch} framework alongside the transformers~\cite{wolf2019huggingface} library from Hugging Face to handle the computational demands of our document re-ranking tasks.

\section{Additional Results}
\subsection{The performance improvement}
\label{apendix:performance-improvement}
The performance improvement of the \model is focused on using zero-shot answer scent generation with a cross-attention mechanism within its re-ranking framework. \model uses the advanced capabilities of LLMs to interpret and generate answer scents. The answer scent is not static but dynamically interacts with the passage tokens through a cross-attention mechanism employed in the model's architecture.  Each token of the generated answer scent considers every token in the passage, enabling a deeper and more contextual understanding before determining the relevance of each passage. By focusing on the semantic and contextual alignment between the question and the document, \model improves the retrieval and ensures that the top-ranked documents are relevant to the information needs.

\subsection{Archival Questions Result}
\label{appendix:archival-result}

In this section, we present the results for the ArchivalQA dataset using different retrievers, both before and after re-ranking with Llama-3-70B and GPT-3.5-turbo-0125 models combined with the T5-base model. The baseline results include performance metrics for Contriever, BM25, DPR, Rocket, and ANCE, showing their Top-1, Top-5, and Top-10 retrieval accuracy and average performance. Among the baselines, BM25 performs the best in Top-1, Top-5, Top-10 metrics, with DPR following closely.

After applying re-ranking with Llama-3-70B, all retrievers see significant improvements. The re-ranked results for DPR and ANCE particularly stand out, both achieving the highest accuracy in Top-1 (27.5\% and 27.3\% respectively) and Top-5/Top-10 metrics. 

The results for re-ranking with GPT-3.5-turbo-0125 similarly show improvements across all metrics. ANCE continues to perform well, achieving the highest Top-1 accuracy (28.1\%) after re-ranking with GPT-3.5-turbo, while DPR and BM25 also show enhanced performance compared to their baseline results. 
\input{tables/tab-archival-questions}

\subsection{Impact of Passage Number on Retrieval Accuracy and Latency}
\label{ref:tab-time-sec}
In this section, we analyze the relationship between the number of passages re-ranked and both retrieval accuracy and latency. This study highlights how the \model performs as we increase the number of passage candidates, focusing on Top-K retrieval accuracy and the time taken per query.
We conducted experiments using the NQ development set to evaluate the performance of \model with different quantities of retrieved passages. The passages were retrieved using BM25 and re-ranked using LLaMA (8B) and GPT (175B) models. We varied the number of passages from 100 to 1000 to observe the impact on Top-K accuracy and latency.
The results of these experiments are presented in Table \ref{tab:qa-time}. The table illustrates how increasing the number of re-ranked passages affects the Top-1, 5, 10 retrieval metrics, and the latency per query.
\input{tables/tab-time-sec}
\paragraph{Retrieval Accuracy: }The Top-1 accuracy significantly improves as the number of re-ranked passages increases. For example, using Llama3 8B, Top-1 accuracy increases from 39.2\% with 100 passages to 40.0\% with 750 passages. Similarly, GPT 175B shows an increase in Top-1 accuracy from 44.9\% with 100 passages to 47.5\% with 1000 passages.
\input{tables/tab-llama_7b_open-domain}
\paragraph{Latency:} As expected, the latency per query increases with the number of passages. With Llama3 8B, the latency grows from 3 seconds for 100 passages to 6.7 seconds for 1000 passages. GPT 175B, while providing better accuracy, also shows an increase in latency, from 0.4 seconds for 100 passages to 3.8 seconds for 1000 passages.

\subsection{Comparative Analysis of LLama 7B and UPR for Document Re-Ranking}
\label{appendix:llama_upr}

In this section, we present a comparison between the performance of \model utilizing the LLama 7B model and the UPR method. This analysis is aimed at understanding how \model, enhanced with the capabilities of LLama 7B, measures up against UPR in terms of improving retrieval accuracy across various question-answering datasets.

We evaluated both LLama 7B with \model and UPR across three major datasets: NQ, TriviaQA, and WebQ. The goal was to assess the improvements in retrieval accuracy, specifically focusing on Top-1, Top-5, and Top-10 metrics. The retrieval setups included unsupervised and supervised retrievers.
The detailed results are summarized in Table \ref{tab-open-domain-rank-llama7b}. The analysis highlights the performance of the two methods under different retrievers, providing insights into their effectiveness across varying retrieval conditions.

\paragraph{Performance across Datasets:} Both methods improve retrieval accuracy across all datasets. However, \model with LLama 7B consistently achieves a higher Top-1 metric compared to UPR, suggesting that the inclusion of the answer scent concept might be more effective at distinguishing the most relevant documents at the top of the retrieval list.

\paragraph{Influence of Retrieval Method:} When combined with MSS, \model with LLama 7B surpasses UPR in Top-1 retrieval accuracy by a notable margin (e.g., 41.3\% vs. 38.7\% on NQ). This indicates that \model's approach to utilizing deep contextual embeddings effectively captures nuances that improve the alignment between the query and retrieved documents.

\subsection{FLOPs and Latency Comparison}

To evaluate the efficiency of our approach, we compute the total FLOPs and latency required for reranking 1,000 passages, comparing our method with the UPR method. The FLOPs were calculated using the fvcore and calflops library~\cite{calflops}.

\textbf{UPR Method:} The total number of FLOPs for reranking 1,000 passages is approximately \(2 \times 10^{15}\) FLOPs.

\textbf{\model:} The total number of FLOPs required is significantly lower, at approximately \(1.1 \times 10^{12} + 1.1 \times 10^{15} \approx 1.1011 \times 10^{15}\) FLOPs. This is because our approach generates the answer scent using a larger model (Llama 8B), but only once for the question. Subsequently, the reranking process is handled by a smaller model, such as T5 base, which uses the generated answer scent to rerank the passages. 

For latency comparisons, we measured the time needed to rerank 1,000 passages based on the computational times of GPT, Llama 3-8B, and UPR T0 3B models. For our method, Llama 3-8B is used once to generate the answer scent in 2.77 seconds, and T5 base requires 2.54 seconds to rerank 1,000 passages, leading to a total of 5.31 seconds. In contrast, UPR T0 3B takes an average of 6.7 seconds to rerank the same number of passages. These results highlight the efficiency of our approach, as it reduces both the computational cost (FLOPs) and the reranking latency compared to UPR.

\section{RAG}
\label{appendix-rag}
In the realm of Retrieval-Augmented Generation (RAG), our study delves into the effects of utilizing LLaMA 7B and LLaMA 13B models, along with varying the number of documents considered in the re-ranking process. Our examination reveals differences in performance across two scenarios: using either one or two documents during the re-ranking phase.

Starting with the LLaMA 7B model, we observed that increasing the number of documents from one to two generally improves the recall and contextual understanding of the model, which is critical in generating accurate responses. For instance, when using the MSS-DPR retriever with LLaMA 7B, the exact match (EM) score sees a slight improvement from 24.3\% with one document to 24.9\% with two documents. This pattern is consistent across other retrievers like BM25 and Contriever, suggesting that the additional context from a second document helps the model refine its answers.

Switching to the LLaMA 13B model, which offers more capacity and potentially finer understanding due to its larger size. For example, when using the BM25 retriever with LLaMA 13B, the EM score increases from 18.5\%  to 28.8\% with two documents. This suggests that the larger model can leverage the extra information more effectively, leading to better overall performance.

\input{tables/tab-rag}

\section{Case Study}
In this section, we present a detailed case study to illustrate the effectiveness of \model in re-ranking documents retrieved by different retrieval systems. Tables \ref{apendix:nq_case_study_retriever}, \ref{apendix:web_case_study_retriever}, and \ref{apendix:trivaqa_case_study_retriever} showcase examples from the NQ dev dataset, WebQA, and TriviaQA, respectively. Each table lists the document IDs retrieved before and after applying \model, indicating whether each document contains the answer (`has\_answer: True` or `False`). These case studies demonstrate how \model enhances the precision of document retrieval across varied contexts and query types by leveraging the answer scent generated from advanced language models.

\label{apendix:case_study}
\begin{table*}[p]
	\small
	\begin{tabular}{@{}p{\textwidth}@{}}
		\toprule                                   \textbf{Answer Scent Prompt:}  Generate a brief, insightful answer scent to the following question: $q$\\ 
        
        \midrule                 
		\textbf{Question:} who sang i just called to say i love you?  \\
		\textbf{Answer Scent:} \textcolor{blue}{The song "I Just Called to Say I Love You" was sung by Stevie Wonder. It was released in 1984 and was featured in the romantic comedy film "The Woman in Red". The song was a huge commercial success, reaching number one on the Billboard Hot 100 chart and winning several awards, including the Academy Award for Best Original Song.}  \\
  \textbf{True Answer:} \textcolor{red}{Stevie Wonder}\\
  
        \textbf{DPR: }  \\
        \texttt{6069735, has\_answer: True} \\
        \texttt{6069737, has\_answer: True} \\
        \texttt{8421731, has\_answer: False} \\
        \texttt{8421729, has\_answer: False} \\
        \texttt{8421736, has\_answer: False} \\
        \textbf{DPR+\model: }  \\
        \texttt{6069735, has\_answer: True} \\
        \texttt{6069737, has\_answer: True} \\
        \texttt{12952376, has\_answer: True} \\
        \texttt{5693954, has\_answer: False} \\
        \texttt{6069736, has\_answer: False} \\
        
  \midrule 
  \textbf{Question:}   who is the coach of arizona state men's basketball? \\
  \textbf{Answer Scent:} \textcolor{blue}{ the head coach of the Arizona State University men's basketball team is Bobby Hurley.}  \\
  \textbf{True Answer:} \textcolor{red}{Bobby Hurley}\\
  \textbf{DPR: }  \\
        \texttt{12295065, has\_answer: True} \\
        \texttt{20172309, has\_answer: False} \\
        \texttt{10047172, has\_answer: False} \\
        \texttt{10047129, has\_answer: False} \\
        \texttt{17545880, has\_answer: False} \\
\textbf{DPR+\model: }  \\
        \texttt{4833080, has\_answer: True} \\
        \texttt{4833089, has\_answer: True} \\
        \texttt{20047916, has\_answer: True } \\
        \texttt{20776179, has\_answer: True} \\
        \texttt{20776180, has\_answer: True } \\

    \midrule 
  \textbf{Question: who played mason on wizards of waverly place?}   \\
  \textbf{Answer Scent:} \textcolor{blue}{The character Mason Greyback, a werewolf and love interest of Alex Russo, was played by actor Gregg Sulkin on the Disney Channel show "Wizards of Waverly Place". }  \\
  \textbf{True Answer:} \textcolor{red}{Sulkin}\\
  \textbf{DPR: }  \\
        \texttt{4309105, has\_answer: False} \\
        \texttt{15243289, has\_answer: False} \\
        \texttt{19878866, has\_answer: False} \\
        \texttt{17449330, has\_answer: False} \\
        \texttt{5763564, has\_answer: False} \\
\textbf{DPR+\model: }  \\
        \texttt{10341922, has\_answer: True} \\
        \texttt{9651553, has\_answer: True} \\
        \texttt{17449330, has\_answer: False} \\
        \texttt{17449350, has\_answer: False} \\
        \texttt{17449363, has\_answer: False} \\
   \midrule 
  \textbf{Question: who plays the queen on game of thrones? }   \\
  \textbf{Answer Scent:} \textcolor{blue}{ The role of Cersei Lannister, the Queen Regent of the Seven Kingdoms, on the hit HBO series Game of Thrones is played by the talented actress Lena Headey.}  \\
  \textbf{True Answer:} \textcolor{red}{Headey}\\
  \textbf{DPR: }  \\
        \texttt{15742943, has\_answer: False} \\
        \texttt{16420109, has\_answer: False} \\
        \texttt{13408263, has\_answer: False} \\
        \texttt{15179481, has\_answer: False} \\
        \texttt{13408259, has\_answer: False} \\
\textbf{DPR+\model: }  \\
        \texttt{4456301, has\_answer: True} \\
        \texttt{5396490, has\_answer: True} \\
        \texttt{4456267, has\_answer: True} \\
        \texttt{5396478, has\_answer: True} \\
        \texttt{12800938, has\_answer: True} \\
              
  \bottomrule             
	\end{tabular}
	\caption{Case study of the retrieved passage from DPR retriever before/after reranking for NQ dev dataset. }
	\label{apendix:nq_case_study_retriever}
\end{table*}

\begin{table*}[p]
	\small
	\begin{tabular}{@{}p{\textwidth}@{}}
		\toprule                                  \textbf{Answer Scent Prompt:}  Generate a brief, insightful answer scent to the following question: $q$\\ 
        
        \midrule                
		\textbf{Question: where is jamarcus russell from?}   \\
		\textbf{Answer Scent:} \textcolor{blue}{JaMarcus Russell, the former NFL quarterback, is from Mobile, Alabama.}  \\
  \textbf{True Answer:} \textcolor{red}{Mobile}\\
  
        \textbf{MSS: }  \\
        \texttt{7564117, has\_answer: False} \\
        \texttt{9565056, has\_answer: False} \\
        \texttt{7564095, has\_answer: False} \\
        \texttt{7564124, has\_answer: True} \\
        \texttt{8298815, has\_answer: False} \\
 
        \textbf{MSS+\model: }  \\
        \texttt{7564126, has\_answer: True} \\
        \texttt{7564096, has\_answer: True} \\
        \texttt{7564097, has\_answer: True} \\
        \texttt{7564125, has\_answer: True} \\
        \texttt{7564127, has\_answer: False} \\

  \midrule 
      
		\textbf{Question: where is the fukushima daiichi nuclear plant located?}   \\
		\textbf{Answer Scent:} \textcolor{blue}{The Fukushima Daiichi Nuclear Power Plant is located in Okuma, Fukushima Prefecture, Japan. It is situated on the Pacific coast of Honshu, the largest island of Japan, approximately 150 miles (240 km) north of Tokyo.}  \\
  \textbf{True Answer:} \textcolor{red}{"Japan", "Okum"}\\
  
        \textbf{MSS: }  \\
        \texttt{20927197, has\_answer: False } \\
        \texttt{3072697, has\_answer: False} \\
        \texttt{20927199, has\_answer: False} \\
        \texttt{10806688, has\_answer: False} \\
        \texttt{18499449, has\_answer: True} \\
 
        \textbf{MSS+\model: }  \\
    \texttt{15436005, has\_answer: False} \\
        \texttt{11148375, has\_answer: True} \\
        \texttt{10483298, has\_answer: True} \\
        \texttt{3541428, has\_answer: False} \\
        \texttt{1682872, has\_answer: True} \\

    \midrule 
      
		\textbf{Question: what does jamaican people speak?}   \\
		\textbf{Answer Scent:} \textcolor{blue}{In Jamaica, the official language is English, which is used in government, education, business, and formal settings. However, the most widely spoken language in Jamaica is Jamaican Patois, also known as Jamaican Creole or Patwa.}  \\
  \textbf{True Answer:} \textcolor{red}{Jamaican English}\\
  
        \textbf{MSS: }  \\
        \texttt{5665719, has\_answer: False} \\
        \texttt{9912963, has\_answer: False } \\
        \texttt{5665720, has\_answer: False} \\
        \texttt{11838832, has\_answer: False} \\
        \texttt{20587290, has\_answer: False} \\
 
        \textbf{MSS+\model: }  \\
        \texttt{4423284, has\_answer: False} \\
        \texttt{1353789, has\_answer: False} \\
        \texttt{8404038, has\_answer: False} \\
        \texttt{4423299, has\_answer: False} \\
        \texttt{4423301, has\_answer: False} \\
   
   \midrule 
      
		\textbf{Question: what is the best sandals resort in st lucia?}   \\
		\textbf{Answer Scent:} \textcolor{blue}{St. Lucia is a beautiful island with several amazing Sandals Resorts to choose from. Each resort has its unique features, amenities, and atmosphere, so the "best" one ultimately depends on your personal preferences and priorities.}  \\
  \textbf{True Answer:} \textcolor{red}{"Micoud Quarter", "Choiseul Quarter", "Praslin Quarter", ..}\\
  
        \textbf{MSS: }  \\
        \texttt{18392196, has\_answer: False} \\
        \texttt{18461202, has\_answer: False} \\
        \texttt{11371584, has\_answer: False} \\
        \texttt{16577459, has\_answer: False} \\
        \texttt{3764126, has\_answer: False} \\
 
        \textbf{MSS+\model: }  \\
    \texttt{5476353, has\_answer: False} \\
        \texttt{18392196, has\_answer: False} \\
        \texttt{3401309, has\_answer: False} \\
        \texttt{3401311, has\_answer: True} \\
        \texttt{6134966, has\_answer: False} \\

  \bottomrule             
	\end{tabular}
	\caption{Case study of the retrieved passage from MSS retriever Before/after Reranking for WebQA. }
	\label{apendix:web_case_study_retriever}
\end{table*}

\begin{table*}[p]
	\small
	\begin{tabular}{@{}p{\textwidth}@{}}
		\toprule                                   \textbf{Answer Scent Prompt:}  Generate a brief, insightful answer scent to the following question: $q$\\ 
        
        \midrule         
		\textbf{Question: which 70s show was based on the british show till death us do part?}   \\
		\textbf{Answer Scent:} \textcolor{blue}{The 1970s show based on the British show "Till Death Us Do Part" is "All in the Family".}  \\
  \textbf{True Answer:} \textcolor{red}{"All In The Family", "Justice For All (TV pilot)", "Stretch Cunningham", ...}\\
  
        \textbf{Contriever: }  \\
        \texttt{9539720, has\_answer: False} \\
        \texttt{6899634, has\_answer: False} \\
        \texttt{475319, has\_answer: False} \\
        \texttt{9549805, has\_answer: False} \\
        \texttt{475315, has\_answer: False} \\
 
        \textbf{Contriever+\model: }  \\
        \texttt{9607452, has\_answer: True } \\
        \texttt{1413988, has\_answer: True } \\
        \texttt{1834891, has\_answer: True } \\
        \texttt{5285410, has\_answer: True} \\
        \texttt{1941863, has\_answer: True} \\

  \midrule 
      
		\textbf{Question: what is the name of terence and shirley conran's dress designer son?}   \\
		\textbf{Answer Scent:} \textcolor{blue}{Jasper Conran!}  \\
  \textbf{True Answer:} \textcolor{red}{"Jaspis", "Bruneau jasper", "Egyptian jasper"}\\
  
        \textbf{Contriever: }  \\
        \texttt{4935862, has\_answer: False} \\
        \texttt{4935861, has\_answer: True} \\
        \texttt{7176709, has\_answer: False} \\
        \texttt{14139592, has\_answer: False} \\
        \texttt{5848573, has\_answer: True} \\
 
        \textbf{Contriever+\model: }  \\
        \texttt{5848571, has\_answer: True} \\
        \texttt{5848575, has\_answer: False} \\
        \texttt{5848577, has\_answer: True} \\
        \texttt{5848576, has\_answer: False} \\
        \texttt{5848573, has\_answer: True} \\

    \midrule 
      
		\textbf{Question: in which country is the sky train rail bridge?}   \\
		\textbf{Answer Scent:} \textcolor{blue}{The SkyTrain Rail Bridge is located in Vancouver, British Columbia, Canada.}  \\
  \textbf{True Answer:} \textcolor{red}{"Canada", "Kenadian", "Canadialand", "Xanada", "Dominion of Canada", "Canadaa"}\\
  
        \textbf{Contriever: }  \\
        \texttt{11617523, has\_answer: False} \\
        \texttt{11617522, has\_answer: False} \\
        \texttt{7697355, has\_answer: False} \\
        \texttt{3375880, has\_answer: False} \\
        \texttt{4904611, has\_answer: True} \\
 
        \textbf{Contriever+\model: }  \\
        \texttt{8509738, has\_answer: True} \\
        \texttt{1145807, has\_answer: True} \\
        \texttt{1145854, has\_answer: True} \\
        \texttt{1145799, has\_answer: True} \\
        \texttt{8509740, has\_answer: True} \\
   
   \midrule 
      
		\textbf{Question: bandar seri begawan international airport is in which country?}   \\
		\textbf{Answer Scent:} \textcolor{blue}{Bandar Seri Begawan International Airport is located in Brunei.}  \\
  \textbf{True Answer:} \textcolor{red}{"Abode of Peace", "BRUNEI", "Health in Brunei", ... }\\
  
        \textbf{Contriever: }  \\
        \texttt{2693267, has\_answer: False} \\
        \texttt{6595413, has\_answer: False} \\
        \texttt{10932719, has\_answer: False} \\
        \texttt{670520, has\_answer: True} \\
        \texttt{10932726, has\_answer: True} \\
 
        \textbf{Contriever+\model: }  \\
        \texttt{670503, has\_answer: True} \\
        \texttt{670496, has\_answer: True} \\
        \texttt{10893158, has\_answer: True} \\
        \texttt{5225731, has\_answer: True} \\
        \texttt{11964123, has\_answer: True} \\

  \bottomrule             
	\end{tabular}
	\caption{Case study of the retrieved passage from Contriever retriever Before/after Reranking for TriviaQA. }
	\label{apendix:trivaqa_case_study_retriever}
\end{table*}

%% file: tables/tab-archival-questions.tex
\begin{table}[h!]
\small
\centering
\begin{adjustbox}{width={0.35\textwidth}}
\begin{tabular}{l | c c c c c }
 \toprule
 \tf{Retriever} & \multicolumn{4}{c}{\tf{ArchivalQA Questions}}  \\
               & Top-1& Top-5& Top-10& Avg \\
\midrule
\multicolumn{5}{c}{\textit{Baselines}} \\
\midrule
Contriever &  1 &3.2 & 5.0 &  3.0\\ 
BM25   & \textbf{18.2} & \textbf{32.3} & \textbf{38.6} &  \textbf{29.7}\\ 
DPR   & 17.0  &30.1  &  36.8& 27.9 \\ 
Rocket & 15.7  & 29.3 & 35.6 &   26.9\\ 
ANCE & 18.0  &31.8 & 37.7 & 29.2 \\ 
\midrule
\multicolumn{5}{c}{\textit{After Re-ranking with Llama-3-70B}} \\
\midrule
Contriever$+T5_{base}$   & 3.9 & 8.1 & 10.4 &  7.4 \\ 
BM25$+T5_{base}$     &  26.2 & 37.3 & 42.4 &   35.3 \\ 
DPR$+T5_{base}$     &\textbf{27.5}  & \textbf{38.2} & \textbf{43.3} &  \textbf{36.3} \\ 
Rocket$+T5_{base}$  &  26.2 & 37.4 & 42.4 &  35.3 \\ 
ANCE$+T5_{base}$  &  27.3 & \textbf{38.2} & \textbf{43.3} &  \textbf{36.3}\\ 
\midrule
\multicolumn{5}{c}{\textit{After Re-ranking with GPT-3.5-turbo-0125}} \\
\midrule
Contriever$+T5_{base}$  & 4.2 &8.7 & 10.9 & 7.9 \\
BM25$+T5_{base}$    & 27.6 & 37.7 & 42.4 & 35.9 \\ 
DPR$+T5_{base}$    & 27.7 &38.5  &43.5 &  36.6\\ 
Rocket$+T5_{base}$    &  26.5 & 37.9 &  42.7 &  35.7 \\ 
ANCE$+T5_{base}$   &  \textbf{28.1} & \textbf{38.1} &\textbf{ 42.9} & \textbf{36.3}\\ 
\bottomrule
\end{tabular}
\end{adjustbox}
\caption{Top-1, 5, 10 retrieval accuracy for the ArchivalQA dataset, comparing baseline retrievers with results after re-ranking using Llama 70b and GPT3.5 models. 
}
\label{tab-archival-questions}
\end{table}

%% file: tables/tab-time-sec.tex
\begin{table}[t]
\small
\centering
\begin{adjustbox}{width=0.5\textwidth}

\begin{tabular}{@{}l | c c c c c c  c c@{}}
 \toprule
 \tf{Retriever /} & \#Document & \multicolumn{5}{c}{\tf{NQ (dev)}}  & Time/Question\\
 \tf{Re-Ranker}         &          &  Top-1 & Top-5 & Top-10 & Top-20 & Top-100  \\
\midrule
BM25   &  -  & 22.3  & 43.8  & 53.7 & 62.3   &  76.0 & - \\  
\midrule
Llama3 8B  & 100  & 39.2 &  58.5 & 64.5 & 69.8 &  76.0 &  3s\\
Llama3 8B  & 200  & 39.6 & 59.4 &  65.7 & 70.9 &  78.5 & 3.4s \\
Llama3 8B  & 250  & 39.7 & 59.5 & 65.8 &71.2  & 79.1  & 3.6s \\
Llama3 8B  & 500  & 39.9 &   60.0 &  66.6  &  71.9 &  80.2 & 4.6s \\
Llama3 8B  & 750  & 40.0 & 60.2 & 66.7  & 71.9 &  80.74 & 5.6s \\
Llama3 8B  & 900  &  39.8 & 60.2 & 66.6 & 72.0 &  80.9 &   6.3s\\
Llama3 8B  & 1000  & 39.8 & 60.2 & 66.6 &  71.9 & 80.9  & 6.7 \\
\midrule
GPT 175B & 100 & 44.9 & 61.1 & 66.3 & 70.8 & 76.0 & 0.4s \\
GPT 175B & 250 & 46.3 & 62.8 & 68.1 &  72.4 & 79.4 & 0.8s \\
GPT 175B & 500 &   46.9 & 63.5 & 69.2 &73.6& 80.8& 1.8s\\
GPT 175B & 750 &   47.2 & 64.2 & 69.3 & 74.2 & 81.6&  2.8s\\
GPT 175B & 900 & 47.4 &  64.4 &69.5  & 74.3 & 81.8 & 3.4s \\
GPT 175B & 1000 & 47.5 & 64.5 & 69.7 & 74.3 & 81.9 & 3.8s \\
\bottomrule
\end{tabular}
\end{adjustbox}
\caption{
Impact of the Number of Passage Candidates on Top-1, Top-5, Top-10 Retrieval Accuracy, and Latency per Query.
}
\label{tab:qa-time}
\end{table}

%% file: tables/tab-llama_7b_open-domain.tex
\begin{table*}[t!]
\addtolength{\tabcolsep}{-0.65pt}
\small
\centering
\begin{adjustbox}{width={0.8\textwidth}}
\begin{tabular}{@{}l | c c c c | c c c c | c c c c  @{}} 
\toprule
\tf{Retriever} & \multicolumn{4}{c}{\tf{NQ }} & \multicolumn{4}{c}{\tf{TriviaQA}} & \multicolumn{4}{c}{\tf{WebQ}}  \\ 
                    & Top-1 & Top-5 & Top-10 & Avg & Top-1 & Top-5 & Top-10 & Avg & Top-1 & Top-5 & Top-10 & Avg \\ 
\midrule
\multicolumn{13}{c}{\textit{Unsupervised Retrievers}} \\
\midrule
MSS               & 19.2 & 41.2 & 51.2 &  37.2 & 30.7 & 52.6  & 60.5   & 47.9 & 11.6  & 29.0 & 39.1 & 26.6  \\ 
MSS + UPR         &  38.7 & \textbf{64.8} & \textbf{72.2} &\textbf{58.6} & 57.2 & \textbf{75.5}  &  78.9 & 70.5 &  29.9 &  57.4 & 65.0 &  50.7 \\ 

MSS + \model$+T5_{base}$       & 41.3   &  60.3 &   67.2 & 56.2 &58.5 & 71.8 & 75.6    & 68.6 & 40.1 & 59.9 & 66.6 &  55.5\\
MSS + \model$+T5_{Large}$       & \textbf{42.8}   & 63.2   & 70.0  & \textbf{58.6}  & \textbf{59.1}  & \textbf{76.4}  & \textbf{79.2}  & \textbf{71.5}  & \textbf{40.4}  & \textbf{60.2}  & \textbf{66.9}  & \textbf{55.8}  \\

\midrule

BM25              & 22.1 & 43.7 & 54.4 &40.1 &  46.3 & 66.2  & 71.7 & 61.4 & 18.8 & 41.8 & 52.1 & 37.6  \\ 
BM25 + UPR        &  35.4 &  63.4 & 70.2& 56.3 & 55.7 & 76.5 & \textbf{80.2}      & 70.8 & 30.1  &57.3  & 66.5 &  51.3\\

BM25 + \model$+T5_{base}$       &  42.1  &  61.1  & 67.4 & 56.8 & 58.2 & 71.1 &  74.7   & 68.0 & 40.9 & 61.1 & 68.1 & 56.7 \\

BM25 + \model$+T5_{Large}$      &  \textbf{44.3}  & \textbf{64.2}  & \textbf{71.0} &  \textbf{59.8} & \textbf{60.8}& \textbf{77.0}&  80.1 & \textbf{72.6} &  \textbf{41.2}& \textbf{62.3} &\textbf{ 68.2} & \textbf{57.2} \\

\midrule
Contriever        &  22.1 & 47.2 & 58.7 & 42.7 & 34.1 &  59.4 &  68.0 & 53.8 &  19.9 &  43.4 & 56.3 & 39.9  \\ 
Contriever + UPR  & 36.4 & \textbf{64.6}  & \textbf{72.4} & \textbf{57.8}&  56.7  & \textbf{77.0}  & \textbf{80.2} & \textbf{71.3} &   30.0 &58.5  & 68.2 &  52.2 \\

Contriever + \model$+T5_{base}$       & \textbf{41.5}  & 61.3  & 68.4 & 57.0 & \textbf{57.9} & 72.8 &  76.8  & 69.1 & \textbf{42.9} & \textbf{62.7} & \textbf{69.8} & \textbf{58.4} \\


\midrule
\multicolumn{13}{c}{\textit{Supervised Retrievers}} \\
\midrule
DPR               &  48.6 & 68.7 & 74.5 & 63.9 & 57.4 &  72.4 &  76.5  & 68.7 & 44.8  & 65.0 & 70.6 & 60.1 \\ 
DPR + UPR         &  42.5 & \textbf{70.6} & \textbf{78.1} & \textbf{63.8} & 61.3 & \textbf{78.7}  &  \textbf{81.9}  & \textbf{74.0} &  34.9 &  63.6 & 71.7 & 56.7  \\ 

DPR + \model$+T5_{base}$        &  \textbf{43.5}  & 64.9  & 72.2 & 60.2 & \textbf{61.8} & 74.6 & 78.3  & 71.5 & \textbf{45.9} & \textbf{66.7} & \textbf{72.4} &  \textbf{61.6}\\ 

\midrule
MSS-DPR           & \textbf{50.1} & \textbf{71.8} & \textbf{77.4} & \textbf{66.5} & 61.6 &  75.2 & 79.1 & 71.9 &  44.2 & 65.0 & 71.6 & 60.3  \\ 
MSS-DPR + UPR     & 41.4 & 69.8 & 77.9 & 63.0 & 60.5 & \textbf{78.9} & \textbf{82.5} & \textbf{74.0} & 31.8  & 61.6 & 70.3 & 54.5  \\ 

MSS-DPR + \model$+T5_{base}$      &  43.5  & 65.1  & 72.5 &60.3  & \textbf{61.7} & 74.8 & 78.6  &  71.7 &  \textbf{44.6} & \textbf{65.4} & \textbf{72.2} & \textbf{60.7} \\

\bottomrule
\end{tabular}
\end{adjustbox}
\caption{
Top-1, 5, 10 retrieval accuracy on the test set of datasets before and after re-ranking the top 1000 retrieved passages.  $\S$ refers to Llama 7B
}
\label{tab-open-domain-rank-llama7b}
\end{table*}

%% file: tables/tab-rag.tex
\begin{table*}[t]
\small
\centering
\begin{adjustbox}{width={0.7\textwidth}}
\begin{tabular}{l | c | c c c | c c c | c c c  }
 \toprule
 \textbf{Model} & top-$K$ & \multicolumn{3}{c}{\textbf{NQ}} & \multicolumn{3}{c}{\textbf{TriviaQA}} & \multicolumn{3}{c}{\textbf{WebQA}}  \\
                &    &  EM & Recall & Con & EM & Recall & Con& EM & Recall & Con \\

\midrule
 \multicolumn{11}{c}{\textbf{{LLama-2 7B}}} \\
 \midrule
Question Only	 & 0 &  14.4 & 28.6 & 21.6 & 41.7 & 54.6 & 48.6 &   14.8 &40.4 &  30.3 \\

\midrule
 \multicolumn{11}{c}{\textbf{{LLama-2 13B}}} \\
 \midrule
Question Only	 & 0 & 11.3 & 27.7 & 20.6 &  39.1 &55.2 & 48.8 & 11.9 &  42.8 & 30.2\\

\midrule
 \multicolumn{11}{c}{\textbf{LLama-2 7B+Baselines}} \\
\midrule
      BM25  & \phantom{00}1     & 16.0 & 29.3 &  21.7 & 51.9  &  63.5 & 57.2 &  14.3 &35.7  & 25.6\\
     
     MSS  & \phantom{00}1     & 14.9 & 27.4 &  20.8 &43.6  & 55.4  & 49.3 &13.7  &37.1  &26.9\\

    Contriever   & \phantom{00}1     &  17.3 &31.1  & 23.9  & 44.4 & 56.5  & 50.2 & 13.6 & 38.5 &23.8 \\

    DPR   & \phantom{00}1     & 24.6 & \textbf{40.5} &  \textbf{32.1} & 50.0 & 62.6  & 56.6 & 15.1 & \textbf{40.3} & \textbf{29.3} \\

    MSS-DPR   & \phantom{00}1     & \textbf{24.9} &  40.4 & 32.0  &   \textbf{51.9} & \textbf{64.7}  & \textbf{58.4} &   \textbf{15.9} &  40.1 & 29.0 \\ 

\midrule
 \multicolumn{11}{c}{\textbf{LLama-2 13B+Baselines}} \\
\midrule
    BM25  & \phantom{00}1     & 18.5 & 30.8 & 23.9 & 54.6 & 65.8  & 59.4 & 14.6  & 35.1 &25.5 \\
    MSS  & \phantom{00}1     & 17.9 & 29.5 & 22.7 &  47.8&  58.5 & 52.6 &  14.9 & 35.6 &25.5 \\
    Contriever   & \phantom{00}1      & 20.3 & 32.4 & 25.2 &  49.0 &  59.5 &53.6 & 17.5  & 38.6 & 28.3 \\
    DPR   & \phantom{00}1     & 27.9 & 43.2 & 33.9 &   53.6 & 65.3  & 58.9 & 18.2  &  \textbf{41.2} & \textbf{30.9} \\
    MSS-DPR   & \phantom{00}1      & \textbf{28.9} & \textbf{43.8} &\textbf{34.5}  &\textbf{54.7} &  \textbf{66.7} & \textbf{60.5} &  \textbf{19.6} & 40.7 & 30.7 \\
\midrule
 \multicolumn{11}{c}{\textbf{LLama-1 7B+Baselines}} \\
\midrule
     MSS  & \phantom{00}2     & 15.1 & 28.2 &  21.2 & 44.6 & 57.3  & 51.3 & 14.3 & 37.8 &27.2 \\
    BM25   & \phantom{00}2 &  16.3 & 29.9 & 22.8  & 52.2 &  65.0 &  58.5 & 13.1 & 36.9 & 26.1 \\
    Contriever   & \phantom{00}2  & 16.9 & 31.3 &  23.9 & 44.8 & 58.2 &51.9 & 13.9 & 39.3 & 28.4 \\
    DPR   & \phantom{00}2 &  23.9 & 39.3  &  31.5 &  49.8 &63.3  &57.0 & 14.8 & \textbf{40.3} & \textbf{28.7}\\
    MSS-DPR   & \phantom{00}2  & \textbf{24.3} & \textbf{40.3} & \textbf{32.2}  & \textbf{50.8} &  \textbf{64.5} &\textbf{58.1}  & \textbf{15.1} &39.7&29.0 \\

    \midrule
     \multicolumn{11}{c}{\textbf{LLama-2 7B+UPR}} \\
\midrule
    MSS  & \phantom{00}1  & 21.9 & 37.3 & 29.2 & 53.5 &  66.5 & 60.1 & 15.1 & 39.9 &\textbf{29.0}\\

     BM25 & \phantom{00}1  & 21.2 &  36.2 & 28.4 & \textbf{57.6} & \textbf{70.3 } & \textbf{63.6 }& 13.9 & 37.0 & 25.6\\

    Contriever & \phantom{00}1  & 22.5 &  38.5 & 30.5 & 53.8 & 67.5 &  61.3& 14.0 & 38.9 & 27.6 \\

    DPR  & \phantom{00}1  & \textbf{23.8} & \textbf{39.8} & \textbf{31.3} & 55.2 &  68.5 & 61.8 & \textbf{15.5} & \textbf{40.1} &28.7\\
    MSS-DPR & \phantom{00}1   & 23.6 &  39.4 & 30.8 &  55.3 & 68.5  & 62.2 &15.3 & 39.8 &28.4\\

    \midrule
         \multicolumn{11}{c}{\textbf{LLama-2 13B+UPR}} \\
\midrule
     MSS  & \phantom{00}1   & 25.2 & 39.4 & 31.2 &  56.4 &  68.4 & 62.1  & 16.9 & 39.3 & 28.0 \\

    BM25 & \phantom{00}1  & 25.1 & 39.3 & 30.7 & 57.3 &  68.8 & \textbf{63.5} & 16.8 & 36.8 &26.7   \\

    Contriever & \phantom{00}1  & 26.0 & 40.4 & 31.9 & 56.5 & 68.0  & 62.7 & 17.4 & 38.3 & 28.4\\

    DPR   & \phantom{00}1   & \textbf{27.4} &  \textbf{42.2} & \textbf{33.0} & \textbf{57.3} & \textbf{69.6} &  63.2 & \textbf{17.5} &\textbf{40.6} & \textbf{29.8}\\

    MSS-DPR   & \phantom{00}1   & 26.3 &41.3 & 32.7 & 57.2  &  69.2 & 62.9 &17.1 & 37.9 &27.2 \\

    \midrule
     \multicolumn{11}{c}{\textbf{LLama-1 7B+UPR}} \\
\midrule
     MSS  & \phantom{00}2     & 21.6 & 37.3 &  29.9 & 54.1 & 67.9  & 61.3 & 15.2 &  39.1 & 28.1\\
     BM25 & \phantom{00}2  & 22.0 & 37.8 & 30.2 &\textbf{ 58.2} &  \textbf{71.4} & \textbf{64.8} & 14.7  & 39.7 & 28.1 \\

    Contriever & \phantom{00}2 & 22.3 & 38.4 & 30.5 & 54.9& 68.2& 65.0  &  14.5 &  38.8 &  27.2  \\

    DPR   & \phantom{00}2  &  23.2 & 38.9 &  31.3 & 55.1 &  69.3 & 62.9 &\textbf{15.7} &\textbf{40.4} &\textbf{28.8}\\
    
    MSS-DPR   & \phantom{00}2   & \textbf{24.1} & \textbf{40.0} &   \textbf{32.0}  &54.9  &  69.2 & 62.7 & 14.4 & 39.6 &27.5\\

    \midrule
    \multicolumn{11}{c}{\textbf{LLama-2 7B+\model}} \\
    \midrule
     MSS  & \phantom{00}1     & 24.8 &  40.6 & 32.7  & 57.1 &  70.5  & 64.1 & 17.9 & 42.3 &  31.2 \\
    BM25   & \phantom{00}1    & 25.0 & 40.4  & 32.3  &   \textbf{60.6}&  \textbf{73.2} & \textbf{66.8} & 16.9 &  42.5 & 31.6 \\
    Contriever   & \phantom{00}1     &  25.9 & 41.9  &  33.6 & 57.8 &  71.0  & 64.9 & 17.7 & 43.9 & 33.1 \\

    DPR   & \phantom{00}1     & 25.8 & 42.2  & 33.7  & 57.6 &  71.1  & 64.6 &  16.5 &  43.7 &  31.2\\

    MSS-DPR   & \phantom{00}1     & \textbf{25.9} & \textbf{42.6 } & \textbf{34.2}  &58.9  & 71.8  &  65.4& \textbf{18.1} & \textbf{43.9} & \textbf{32.8} \\
    \midrule
\multicolumn{11}{c}{\textbf{LLama-2 13B+\model}} \\
    \midrule
     MSS  & \phantom{00}1     & 28.5 &  43.4 &34.6 &60.1 & 72.4  &65.9  &  20.5 &43.9  & 33.6  \\
    BM25   & \phantom{00}1    &  28.8 &  44.2 & 35.4 &\textbf{63.3} & \textbf{74.9}  &  \textbf{68.5} &19.3 & 43.0 & 31.5  \\

    Contriever   & \phantom{00}1     &  29.7 & 45.1 & 36.3 &  60.1 &  72.4 & 66.1 & 20.6 &   44.0 &   33.2 \\

    DPR   & \phantom{00}1     &   28.9 & 44.9  & 35.5 &  60.8 &  72.9 &66.7  & 19.9 & 43.0 &  32.2 \\

    MSS-DPR   & \phantom{00}1     &\textbf{30.2} &  \textbf{45.5}  & \textbf{36.6} &60.9 & 73.3  & 66.9 & \textbf{20.6} &  \textbf{44.7} &  \textbf{33.7} \\

    \midrule
    \multicolumn{11}{c}{\textbf{LLama-1 7B+\model}} \\
\midrule
     MSS  & \phantom{00}2     & 25.5 &  41.3 & 33.2  & 56.5 &  70.8  & 64.2 & 17.9 & 43.8 & 32.9 \\
    BM25   & \phantom{00}2 & 24.9 & 40.8 & 33.4  & \textbf{59.8} & \textbf{73.3}  &  \textbf{67.1} &\textbf{16.6}& \textbf{41.7} & \textbf{30.4}\\
    Contriever   & \phantom{00}2  & 25.5 & 41.6 &  33.8 & 56.5 & 71.4  & 64.9 &  17.0& 43.4 &31.8 \\
    DPR   & \phantom{00}2 &  25.9 & 41.9 & 33.8  &  57.3 &   71.2 & 64.7 & 17.6 & 43.6 &32.3 \\
    MSS-DPR   & \phantom{00}2  & \textbf{26.0} &\textbf{42.3}  &  \textbf{34.4 }&  57.7 & 72.1  & 65.5 & 17.6 & 43.7 & 32.5\\
 \bottomrule
\end{tabular}
\end{adjustbox}
\caption{Exact match scores for the open-domain QA task.
}
\label{tab-rag}
\end{table*}